\documentclass[sigconf,nonacm]{acmart}
\AtBeginDocument{%
  }

\setcopyright{acmlicensed}
\copyrightyear{2025}
\acmYear{2025}

\newcommand{\ourmethod}{PAFT\,}
\newtheorem{definition}{\textbf{Definition}}
\usepackage{amsmath}
\usepackage{soul}
\usepackage{subcaption}
\usepackage{booktabs}
\usepackage{makecell}
\usepackage{multirow}
\usepackage{algorithm}
\usepackage[noend]{algpseudocode}
\usepackage{cleveref}
\usepackage{xcolor}
\usepackage{hyperref}

\begin{document}

\title{Why LLMs Are Bad at Synthetic Table Generation\\ (and what to do about it)}

\author{Shengzhe Xu}
\affiliation{%
  \institution{Computer Science \\Virginia Tech}
  \city{Arlington}
  \state{VA}
  \country{USA}
}

\author{Cho Ting Lee}
\affiliation{%
  \institution{Computer Science \\Virginia Tech}
  \city{Arlington}
  \state{VA}
  \country{USA}
}

\author{Mandar Sharma}
\affiliation{%
  \institution{Computer Science \\Virginia Tech}
  \city{Arlington}
  \state{VA}
  \country{USA}
}

\author{Raquib Bin Yousuf}
\affiliation{%
  \institution{Computer Science \\Virginia Tech}
  \city{Arlington}
  \state{VA}
  \country{USA}
}

\author{Nikhil Muralidhar}
\affiliation{%
  \institution{Computer Science \\Stevens Institute of Technology}
  \city{Hoboken}
  \state{NJ}
  \country{USA}
}

\author{Naren Ramakrishnan}
\affiliation{%
  \institution{Computer Science \\Virginia Tech}
  \city{Arlington}
  \state{VA}
  \country{USA}
}

\renewcommand{\shortauthors}{Xu et al.}


\begin{abstract}
Synthetic data generation is integral to ML pipelines, e.g., to augment training data, replace sensitive information, and even to power advanced platforms like DeepSeek. While LLMs fine-tuned for synthetic data generation are gaining traction, synthetic table generation—a critical data type in business and science—remains under-explored compared to text and image synthesis. This paper shows that LLMs, whether used as-is or after traditional fine-tuning, are inadequate for generating synthetic tables. Their autoregressive nature, combined with random order permutation during fine-tuning, hampers the modeling of functional dependencies and prevents capturing conditional mixtures of distributions essential for real-world constraints. We demonstrate that making LLMs permutation-aware can mitigate these issues.
Our code and data are anonymously hosted\footnote{\url
{https://github.com/ShengzheXu/Permutation-aided-Fine-tuning}}.
\end{abstract}

\maketitle

\section{Introduction} 
Large language models (LLMs) have found applicability in a rich diversity of domains, well beyond their original roots~\cite{achiam2023gpt,brown2020language,miller2009input,chatgpt,touvron2023llama,raffel:20,romera2023mathematical}. 
As so-called foundation models~\cite{liang2022holistic}
they have been shown to be re-targetable to a variety of downstream tasks. Our focus here is to view LLMs as raw synthetic tabular data generators rather than as supporting an analysis or discovery task. Arguably, LLMs are adept at synthetic generation of text, images, videos, code, documentation, and many other modalities. 
The use of LLMs to generate 
tabular data is
quite understudied.

Such synthetic tabular
data generation is integral to 
ML pipelines,
e.g., too augment training data, replace sensitive information, and even to power
advanced platforms like DeepSeek~\cite{liu2024deepseek,guo2025deepseek}.
However, the unique
characteristics
of tabular data manifest as challenges in an LLM-based generative context. 
The most popular incarnations of language models
are auto-regressive models, e.g., LLama~\cite{touvron2023llama}, GPT-x~\cite{chatgpt}, DeepSeek~\cite{liu2024deepseek,guo2025deepseek},
wherein each word or token is generated conditional on past tokens in a sequential manner
using attention models.
In a synthetic data context, each `sentence' typically represents a row of tabular data, and each `word' corresponds to an attribute in that row. 
The previous state-of-the-art models
(GReaT~\cite{borisov2022language}), has advocated the use of
random feature orders 
but as we show in Table~\ref{tab:sec1_motivative_example}, when fine-tuning is done with random feature orders, key relationships are often not captured or, worse, violated.
In particular,
with tabular data,
there are numerous functional dependencies at play and as a result, generating tokens in random orders is bound to cause violations.
There is thus an
{\bf`impedance mismatch' between autogressive LLMs and synthetic data generation.}

The main contributions of this paper are:
\begin{itemize}
    \item We highlight an important deficiency with using LLMs for synthetic tabular data generation and explore the performance of many state-of-the-art generation models in the context of composite and multi-category tabular schema.
    \item We inject knowledge of pre-existing functional relationships among columns into the autoregressive generation process, so that the generated synthetic data respects more real constraints. In particular,
we present a taxonomy of functional dependencies (FDs) whose discovery and organization into a column dependence graph supports their incorporation into the LLM fine-tuning process via a permutation function, leading to our approach dubbed Permutation-aided Fine-tuning.(\ourmethod).
\item We evaluate the performance of \ourmethod on a range of datasets featuring a diverse mix of attribute types, functional dependencies, and complex relationships. Our results demonstrate that \ourmethod is the state-of-the-art in reproducing underlying relationships in generated synthetic data. 
\item Finally, we demonstrate through rigorous experiments that relying just on standard univariate distribution, bivariate correlation, and even the evaluation of downstream machine learning models (which primarily focuses on predicting a single column in a dataset) is grossly insufficient for assessing the quality of synthetic data and propose systematic remedies like measuring violation rates of known domain rules.

\end{itemize}

\section{Related Work}
\begin{table*}[ht!]
\centering
\caption{Comparison of multiple synthetic data generation approaches. The second row showcases the state of West Virginia (WV), which is a subset of the whole data. In the figures, the solid line represents the official border of West Virginia and the \textcolor{blue}{Blue} and \textcolor{orange}{Orange} colors indicate the legal and illegal samples in the synthetic
data respectively. 
}
\label{tab:sec1_motivative_example}
\setlength{\tabcolsep}{0.2mm}
\begin{tabular}{c|cccccc}
\toprule
\rotatebox[origin=l]{90}{Whole Data} 
& \includegraphics[width=0.16\textwidth,height=2.7cm]{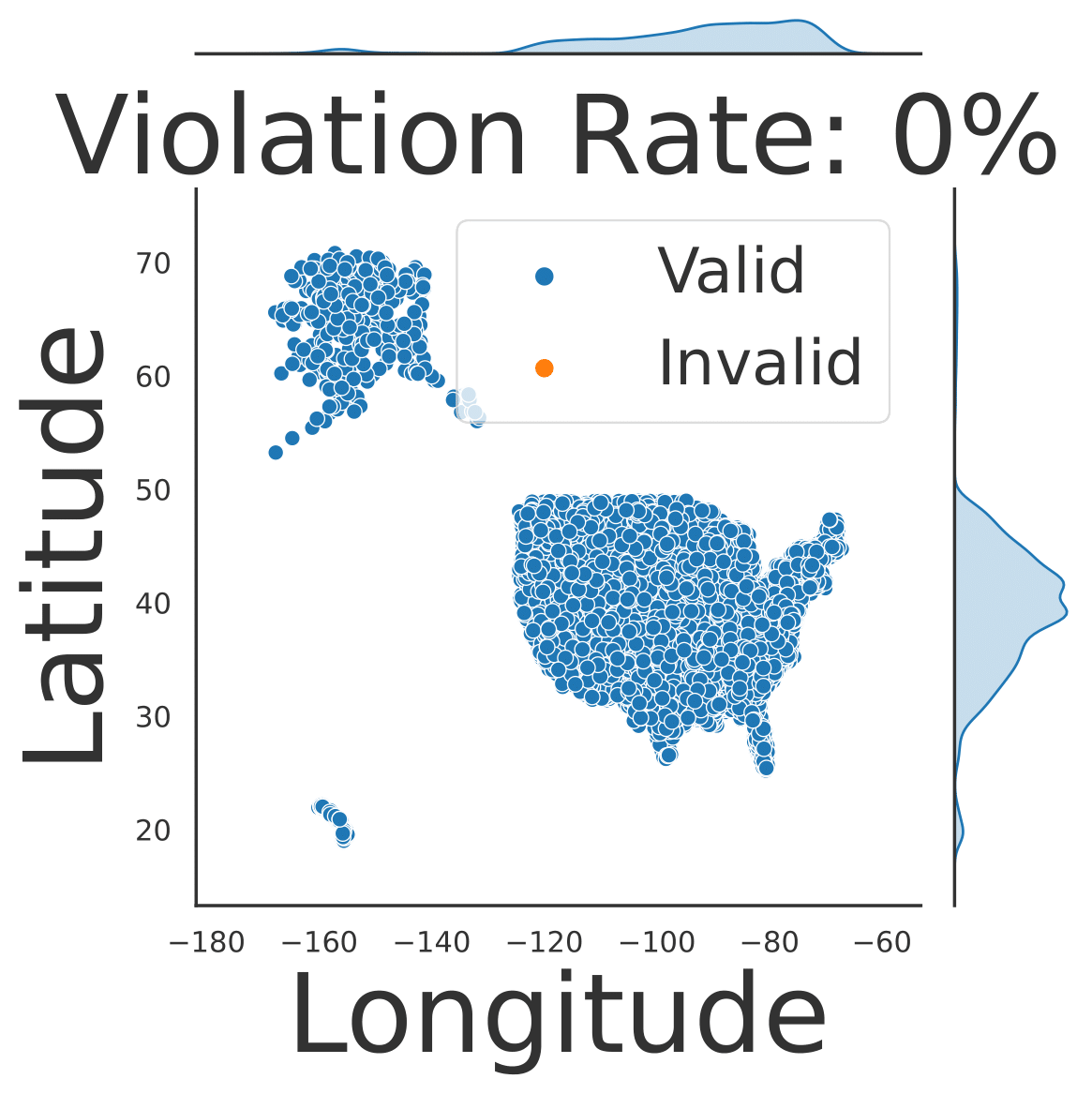} 
& \includegraphics[width=0.16\textwidth,height=2.7cm]{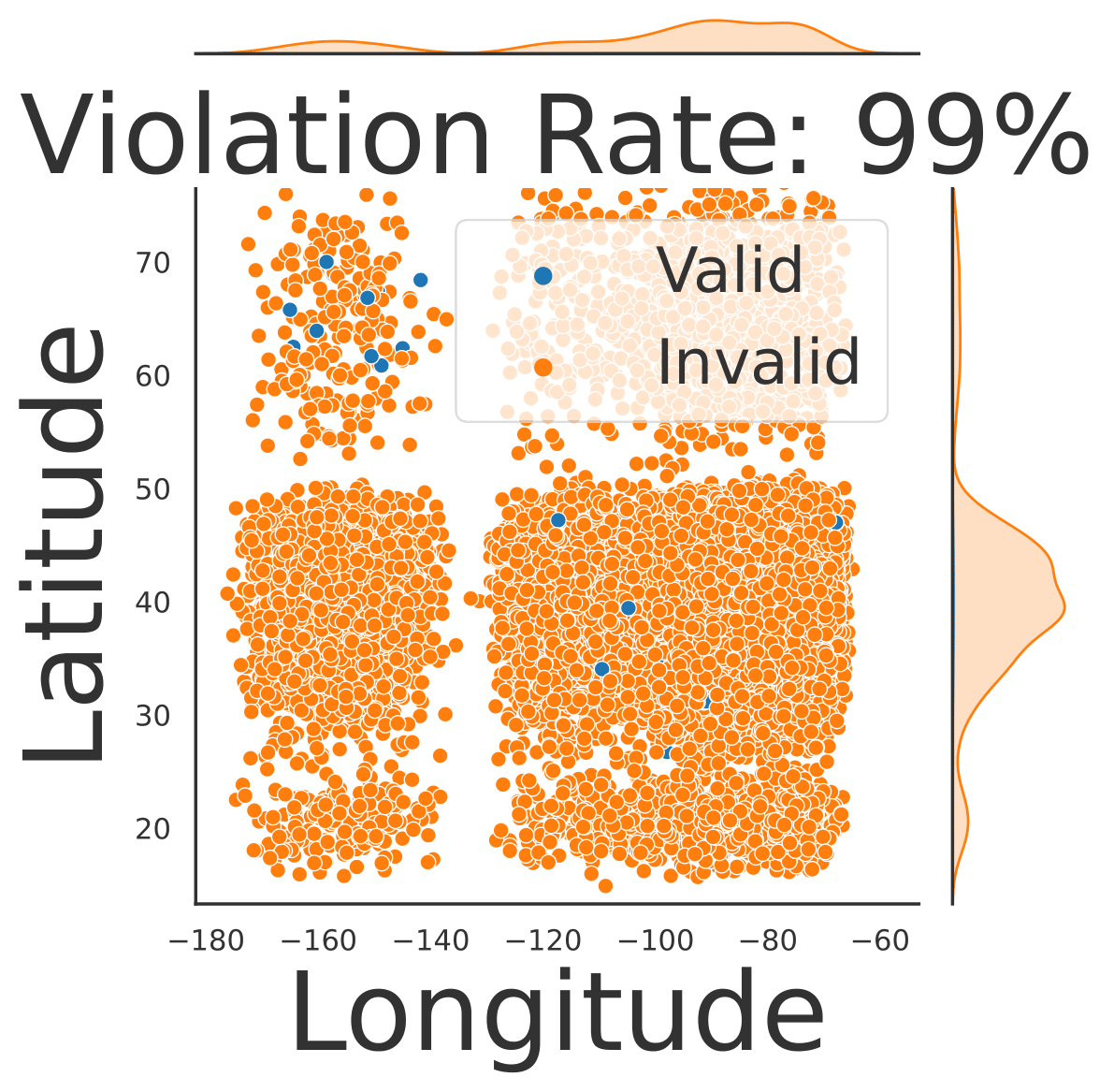} 
& \includegraphics[width=0.16\textwidth,height=2.7cm]{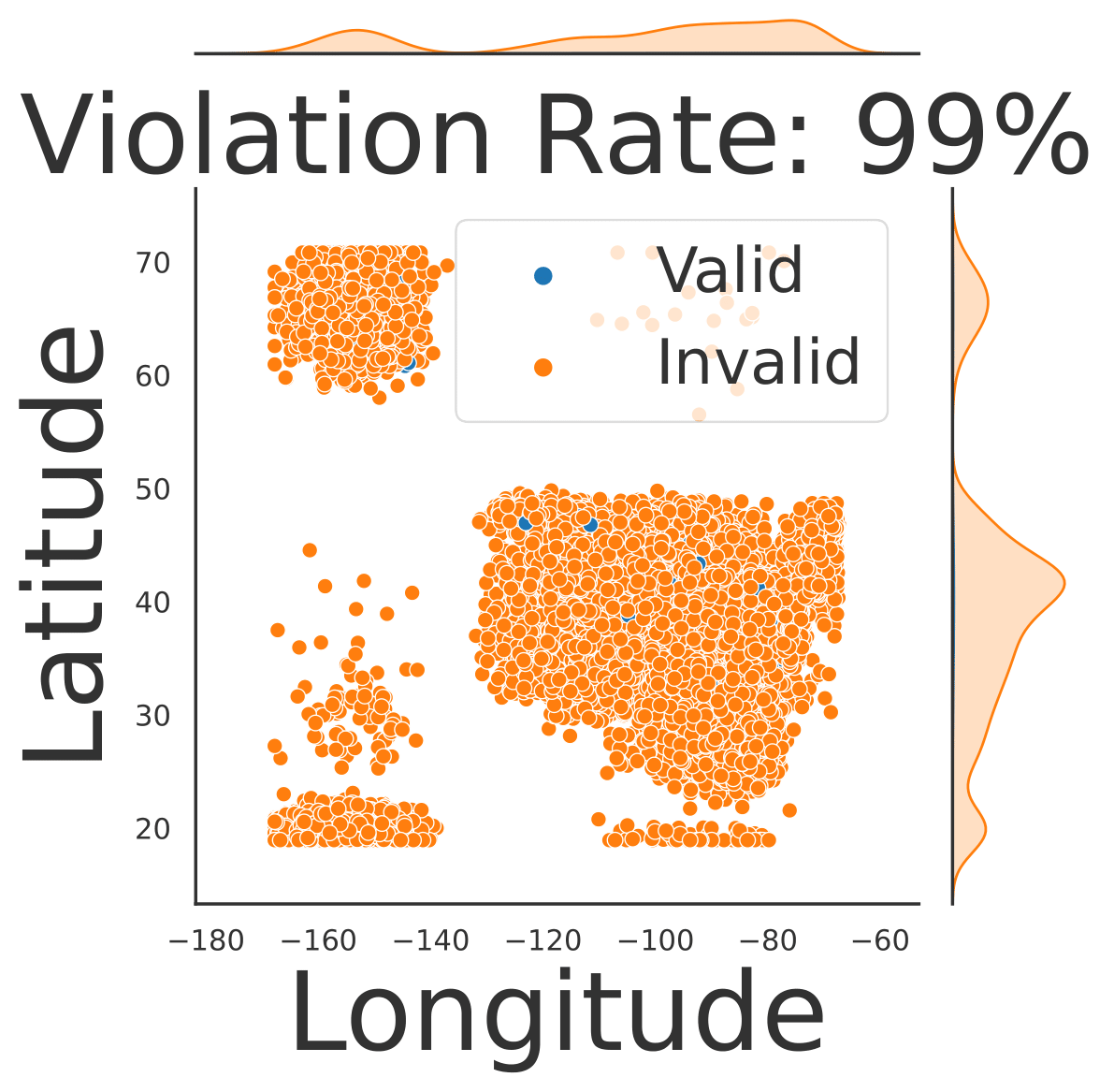}
& \includegraphics[width=0.16\textwidth,height=2.7cm]{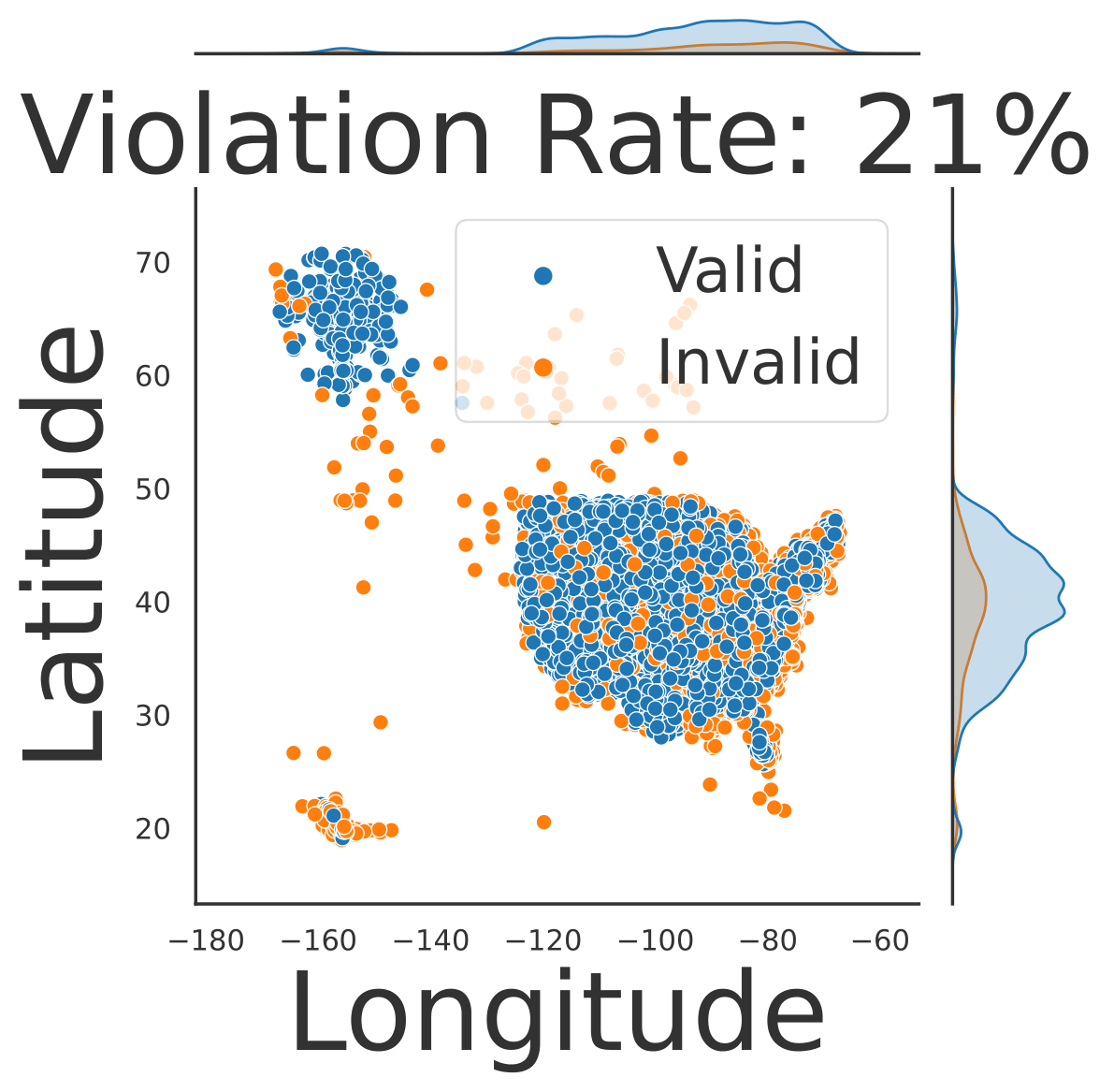}
& \includegraphics[width=0.16\textwidth,height=2.7cm]{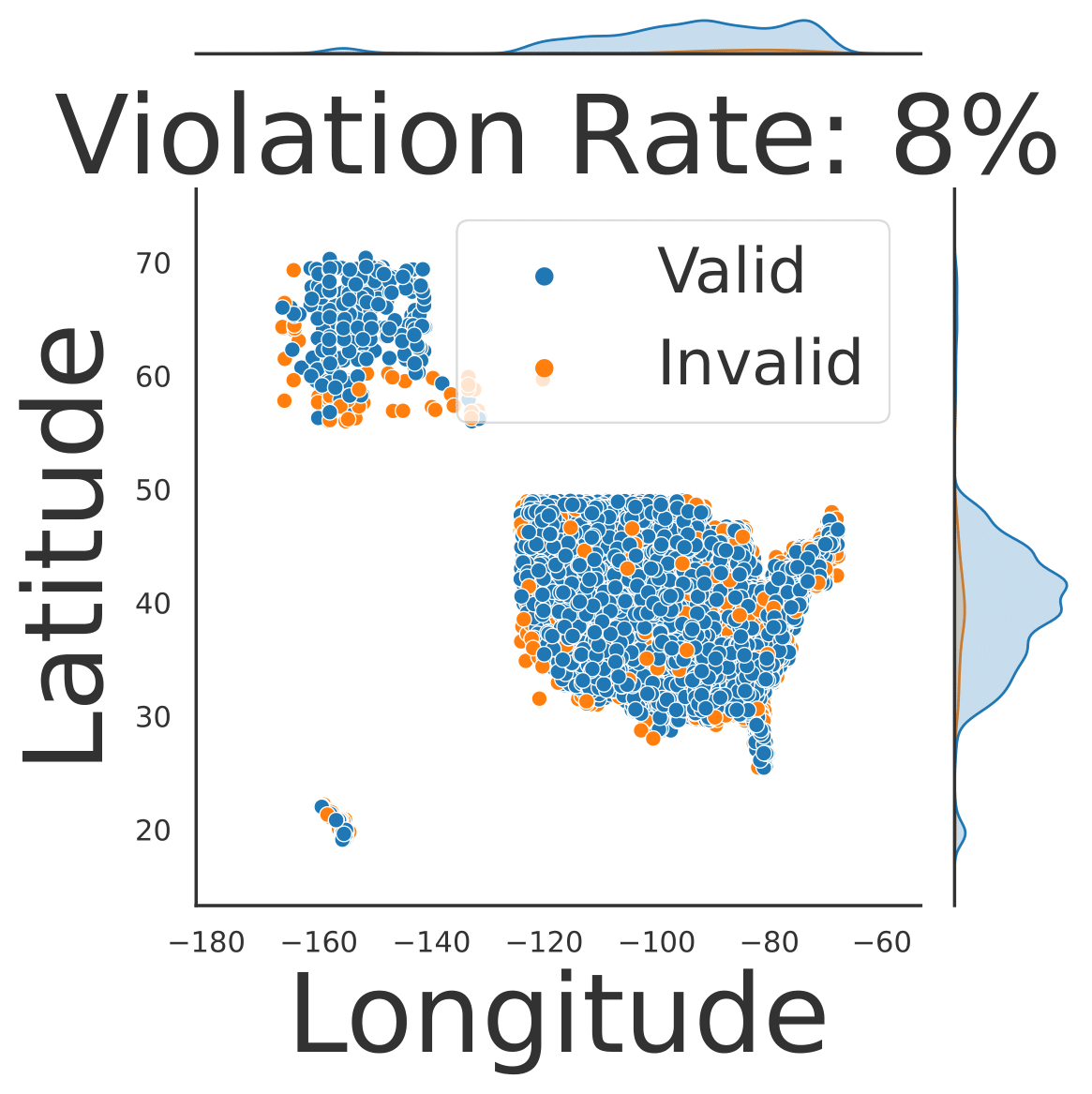}
& \includegraphics[width=0.16\textwidth,height=2.7cm]{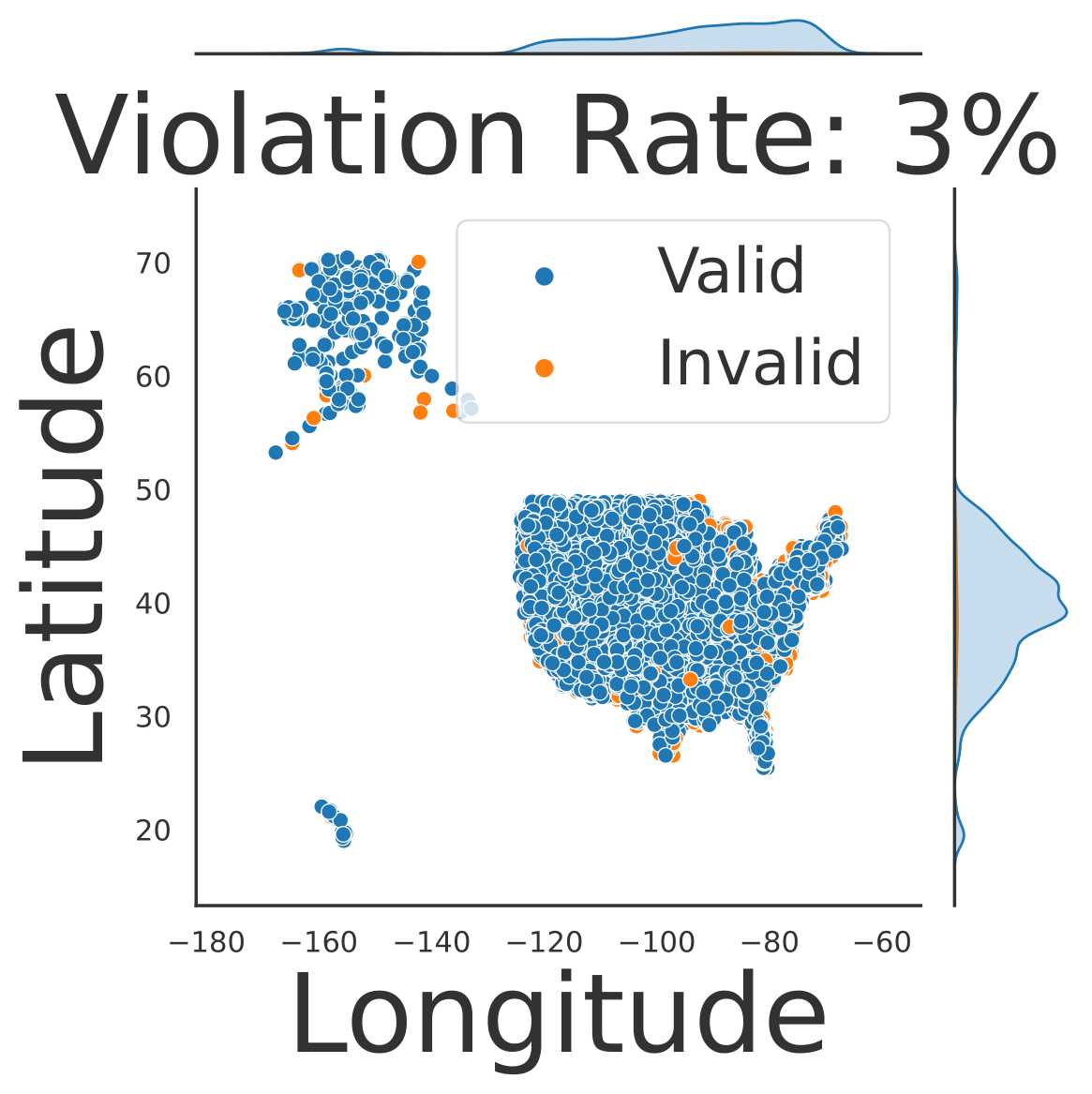}\\
\midrule
\rotatebox[origin=l]{90}{Sub-Category} 
& \includegraphics[width=0.16\textwidth,height=2.7cm]{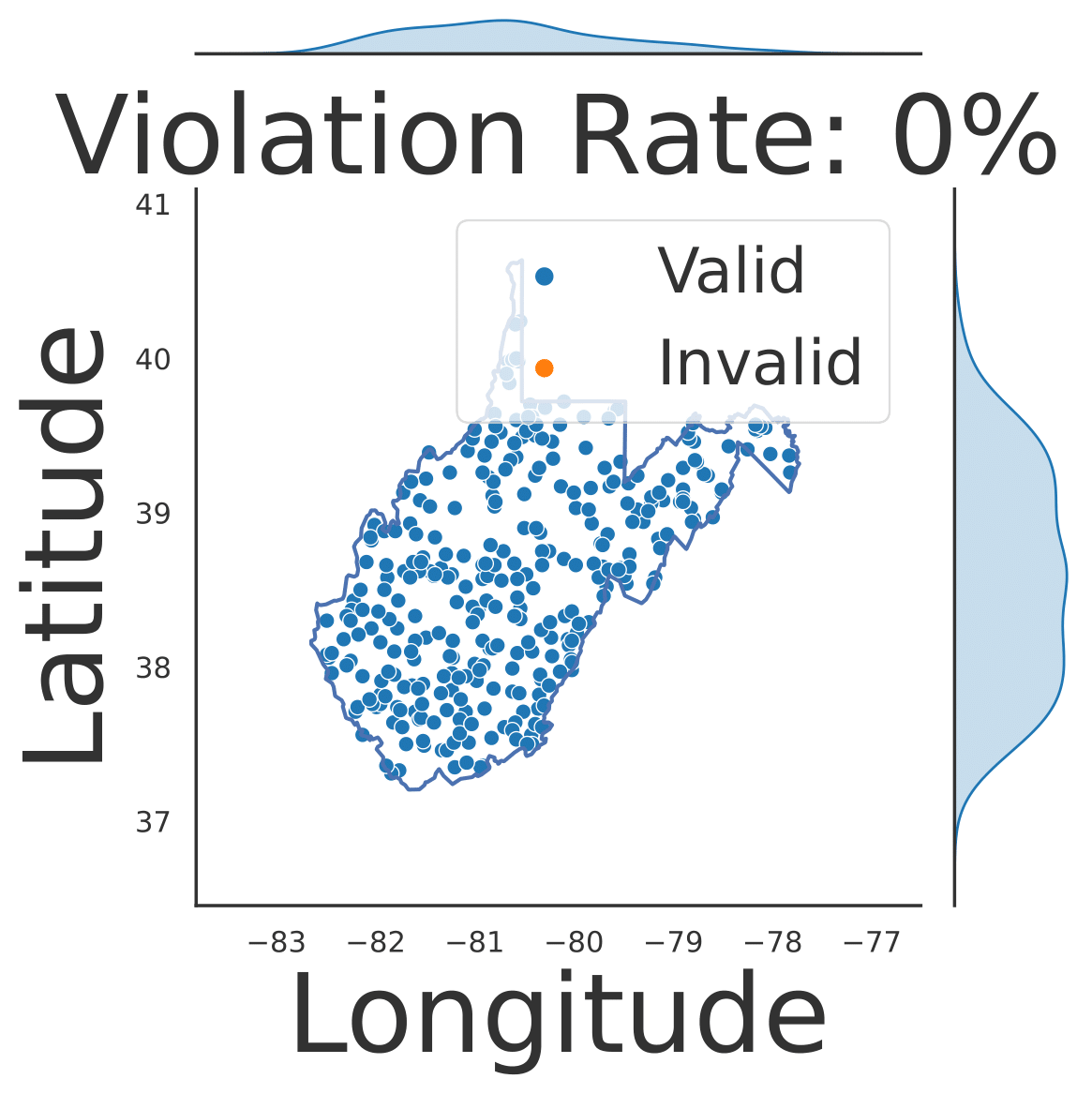} 
& \includegraphics[width=0.16\textwidth,height=2.7cm]{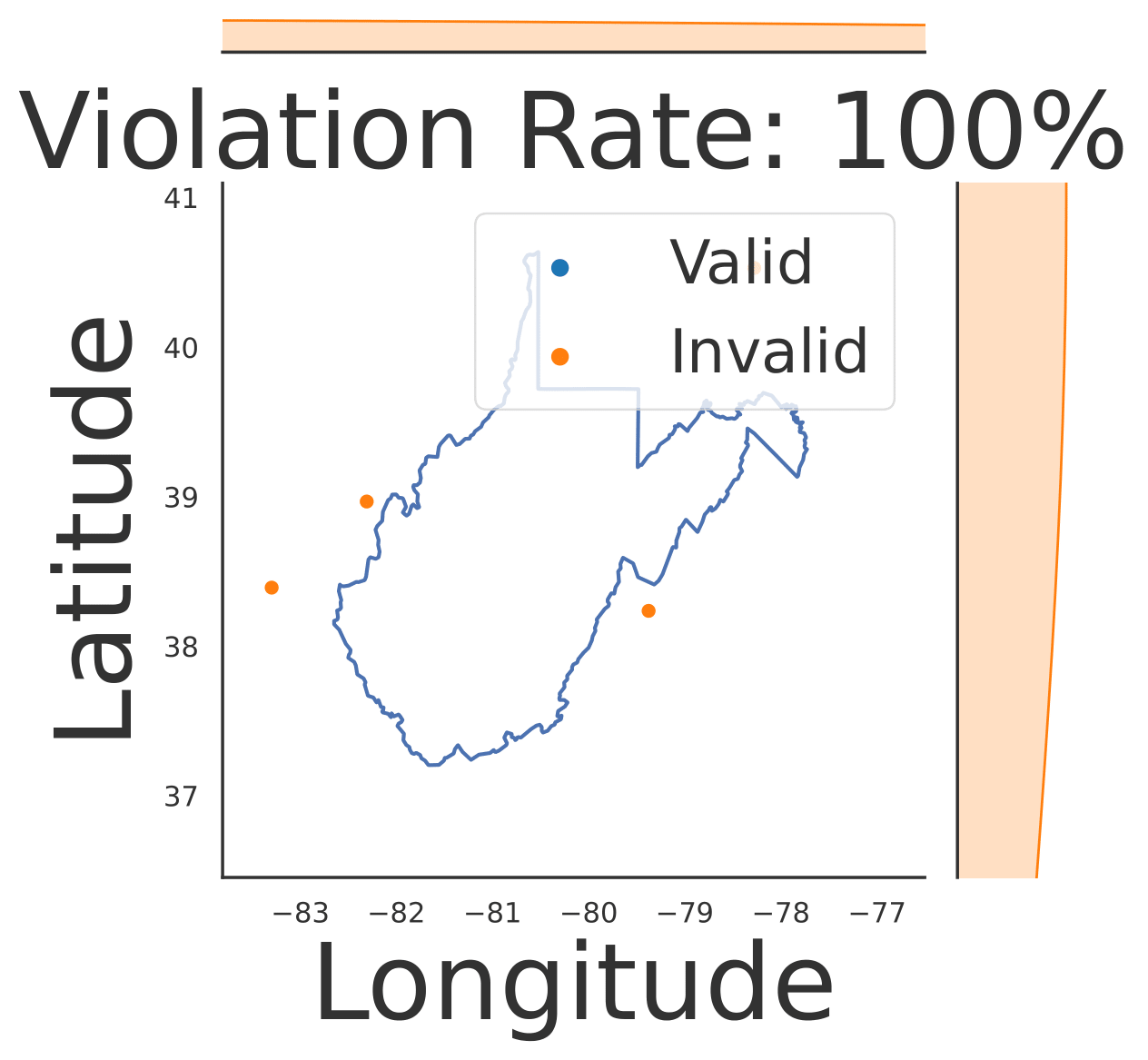} 
& \includegraphics[width=0.16\textwidth,height=2.7cm]{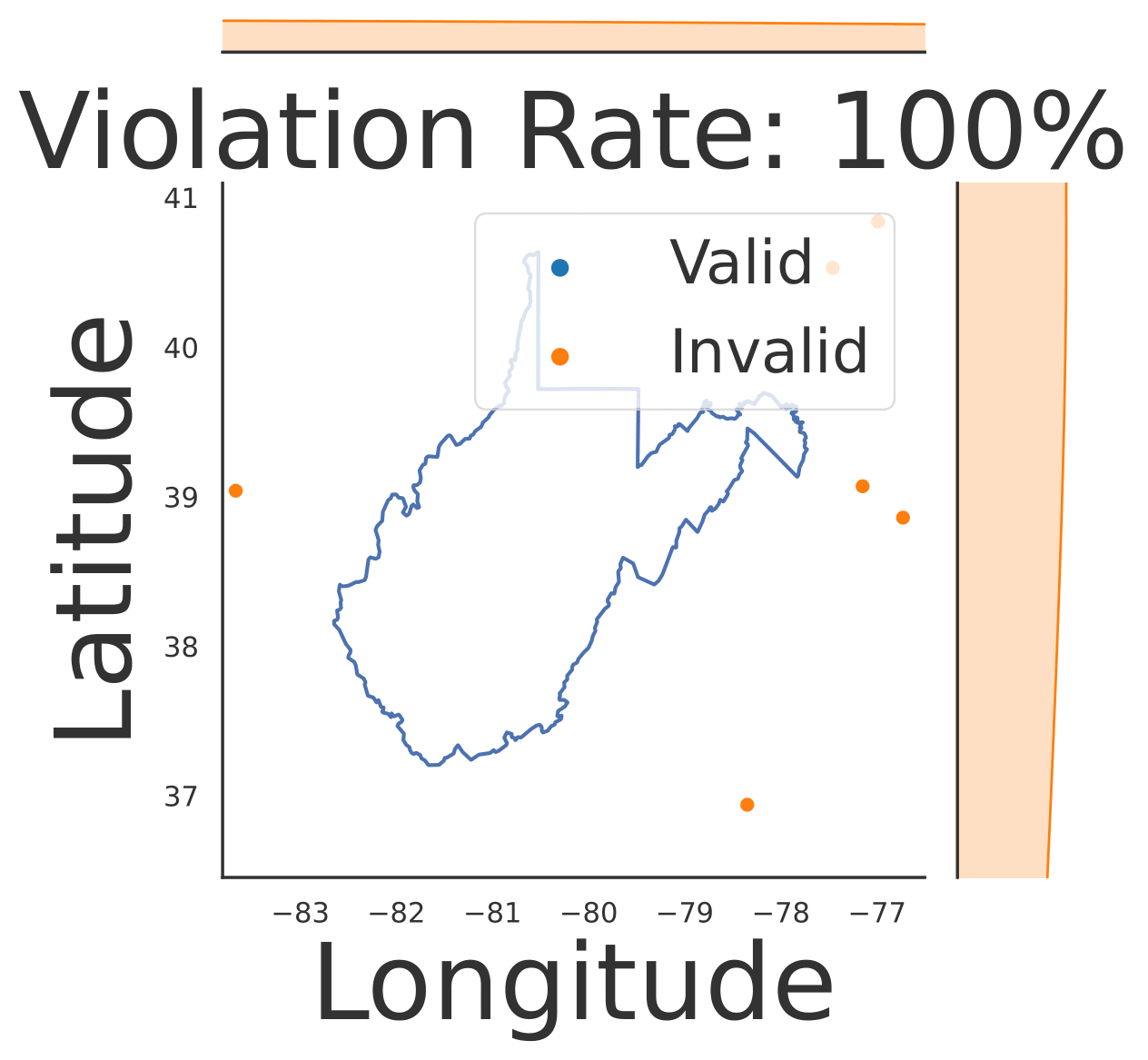}
& \includegraphics[width=0.16\textwidth,height=2.7cm]{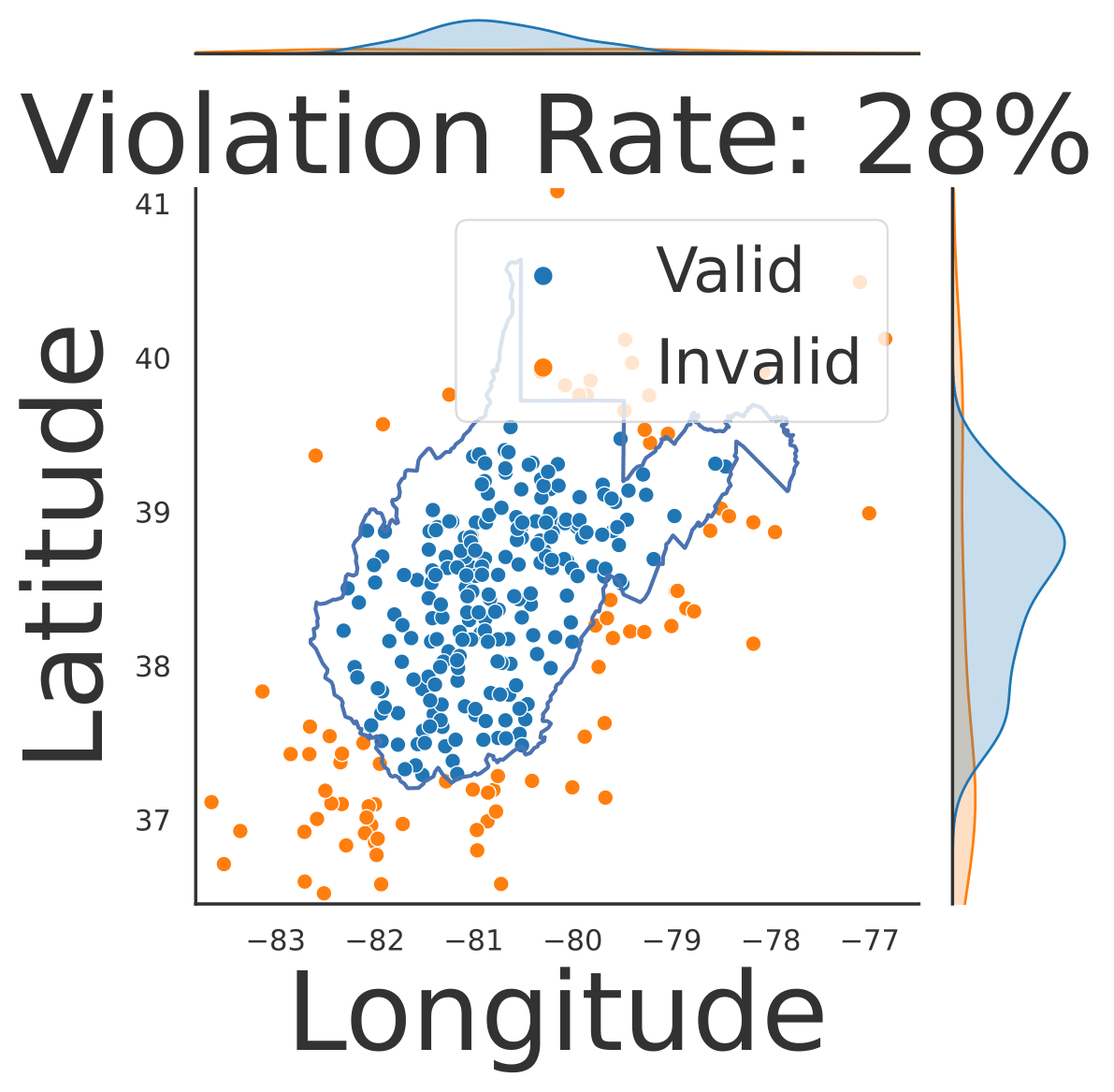}
& \includegraphics[width=0.16\textwidth,height=2.7cm]{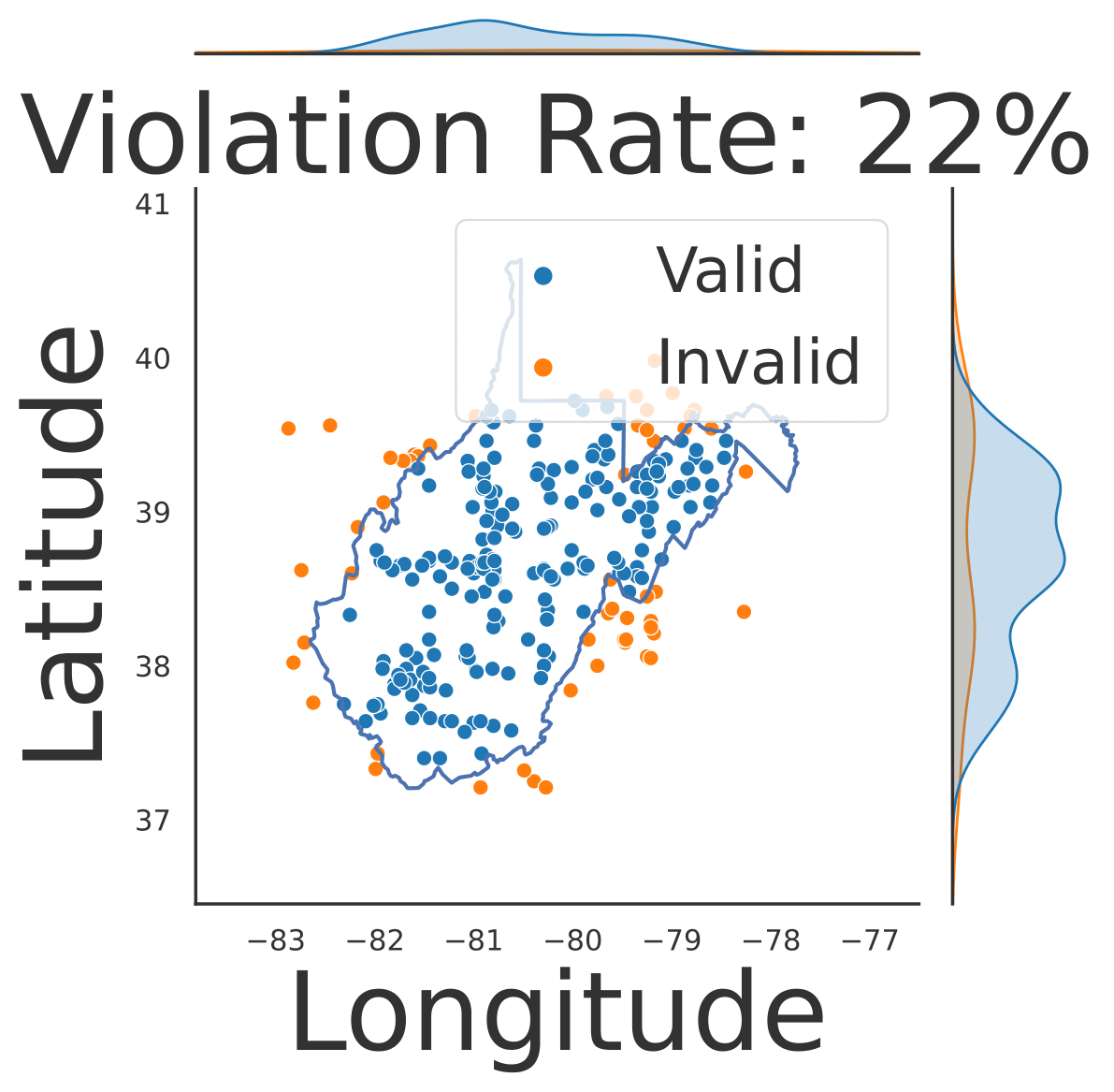}
& \includegraphics[width=0.16\textwidth,height=2.7cm]{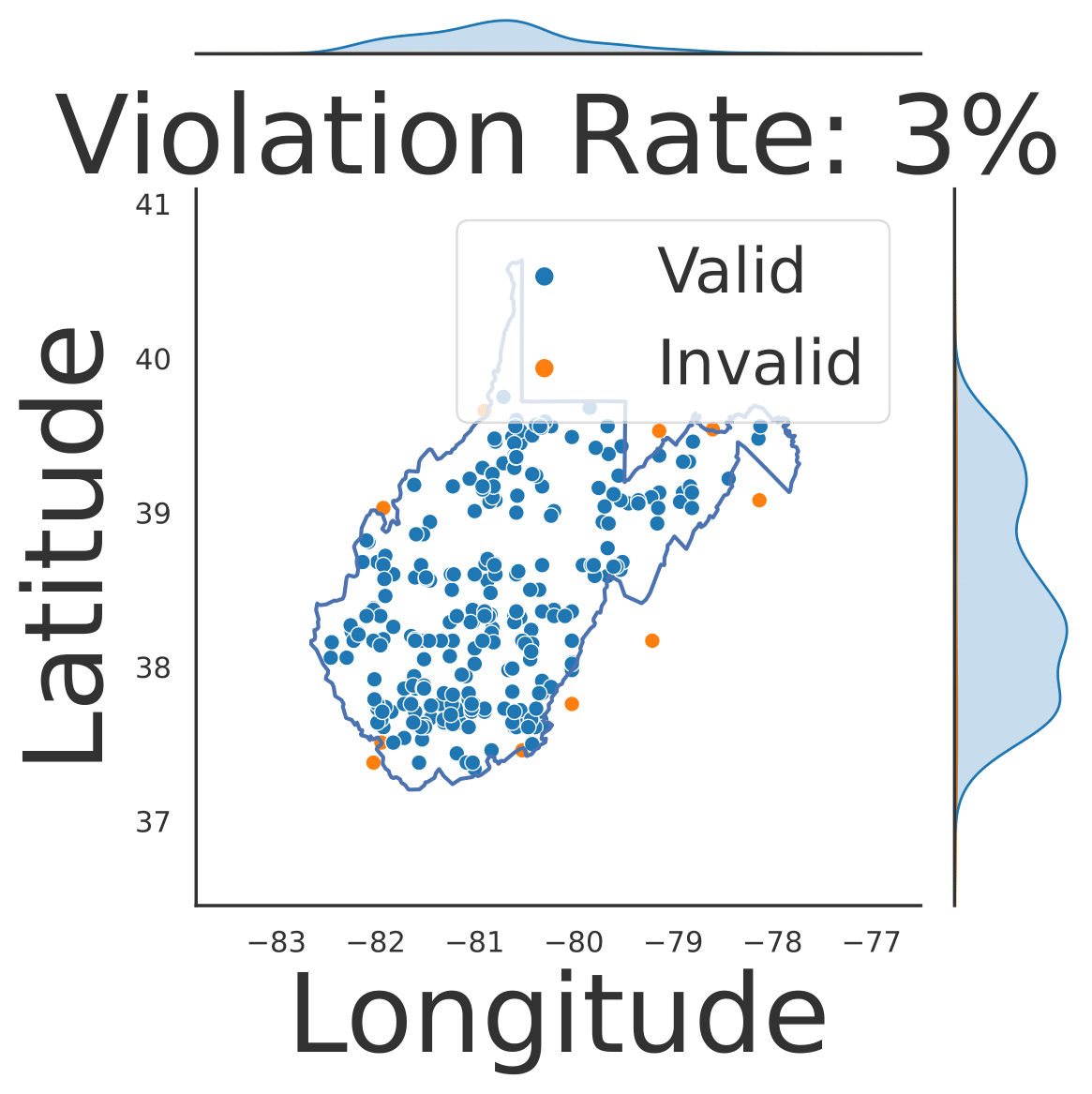}\\
\midrule
& Real & CTGAN & CopulaGAN & TabSyn (Diffusion) & GReaT (LLM-GPT) & PAFT (LLM-GPT) \\
\bottomrule
\end{tabular}
\end{table*}
Tabular data synthesis and representation learning for tables have been
extensively studied~\cite{du2021tabularnet, wang2021tuta,zhang2023mixed,margeloiutabebm,fang2024large,SDV,du2024towards,mckenna2019graphical,park2018data,8805442}. One key theme has been the
discovery of functional
dependencies (FDs) which has been
studied by the data mining
community from both
theoretical and application
points of view~\cite{zhang2020statistical,chen2019faketables,muralidhar2018illiad,mandros2020discovering,pennerath2020discovering,zheng2023dense}.
We carve out related
work into two sections,
for our purposes: pre-LLM (or non-LLM) approaches for synthetic table generation which continue to hold the mainstay, and LLM approaches.

Lei et al.~\cite{xu2019modeling} proposed CTGAN where
rows are independent of each other; a conditional GAN architecture ensures that the dependency between  columns is learned.
Tabsyn~\cite{zhang2023mixed} 
showcased remarkable advancements in joint-distribution learning via a VAE plus diffusion approach, surpassing previous models of similar lineage, in terms of distributional correlation measures and machine learning efficiency.
DoppelGanger~\cite{lin2020using} uses a combination of an RNN and a GAN to incorporate temporal dependencies across rows but this method has been tested in traditional, low-volume settings such as Wikipedia daily visit counts.
For high-volume applications, STAN~\cite{xu2021stan} utilizes a combination of a CNN and Gaussian mixture neural networks
to generate synthetic network traffic data.
GraphDF ~\cite{chen2023graph} conducts multi-dimensional time series forecasting.
GOGGLE ~\cite{liu2022goggle} employs a generative modeling method for tabular data by learning relational structures. 

The use of language models (LLMs) for tabular data generation is still underexplored. Most modern LLMs are based on the transformer architecture ~\cite{attention_all_you_need} with parameters ranging from few millions to billions ~\cite{chinchilla_scaling_law}, and researchers have developed creative ways to harness LLMs in traditional machine learning and data contexts. LIFT~\cite{dinh2022lift} initially transforms a table row into a sentence, such as `An Iris plant with sepal length 5.1cm, sepal width 3.5cm', and employs an LLM as a learning model for table classification, regression, and generation tasks.  GReaT~\cite{borisov2022language} utilizes a GPT-2 model that has been fine-tuned using a specific corpus for synthetic data generation.  They also show that even small-scale models such as Distill-GPT~\cite{radford2019language} have the potential for synthetic data generation~\cite{borisov2022language}. These models are specially viable for tabular generation given the lower compute costs of aligning smaller models to large and varied tabular datasets.  A general benefit of utilizing LLMs is the promise of eliminating customized preprocessing pipelines. 

A theme that will be pertinent to our work is the idea of
feature ordering (FO) for
tabular data generation which has been investigated from multiple angles~\cite{dtg_survey,gtr_lstm,dual_graph,zhu2022permutation}.
There are also several approaches (e.g.,~\cite{kanter2015deep, cvitkovic2020supervised}) that synthesize, discover, or aggregate features from relational databases, leveraging order information when possible, for use in machine learning pipelines.
It worth noting that even in the LLM community, the task of context sorting for LLM prompting is not trivial and has gained significant attention lately~\cite{chen2024premise}.

\section{Challenges to Synthetic Table Generation in the Current LLM Paradigm}

\begin{figure*}[t!]
    \centering
    \includegraphics[width=.75\linewidth]{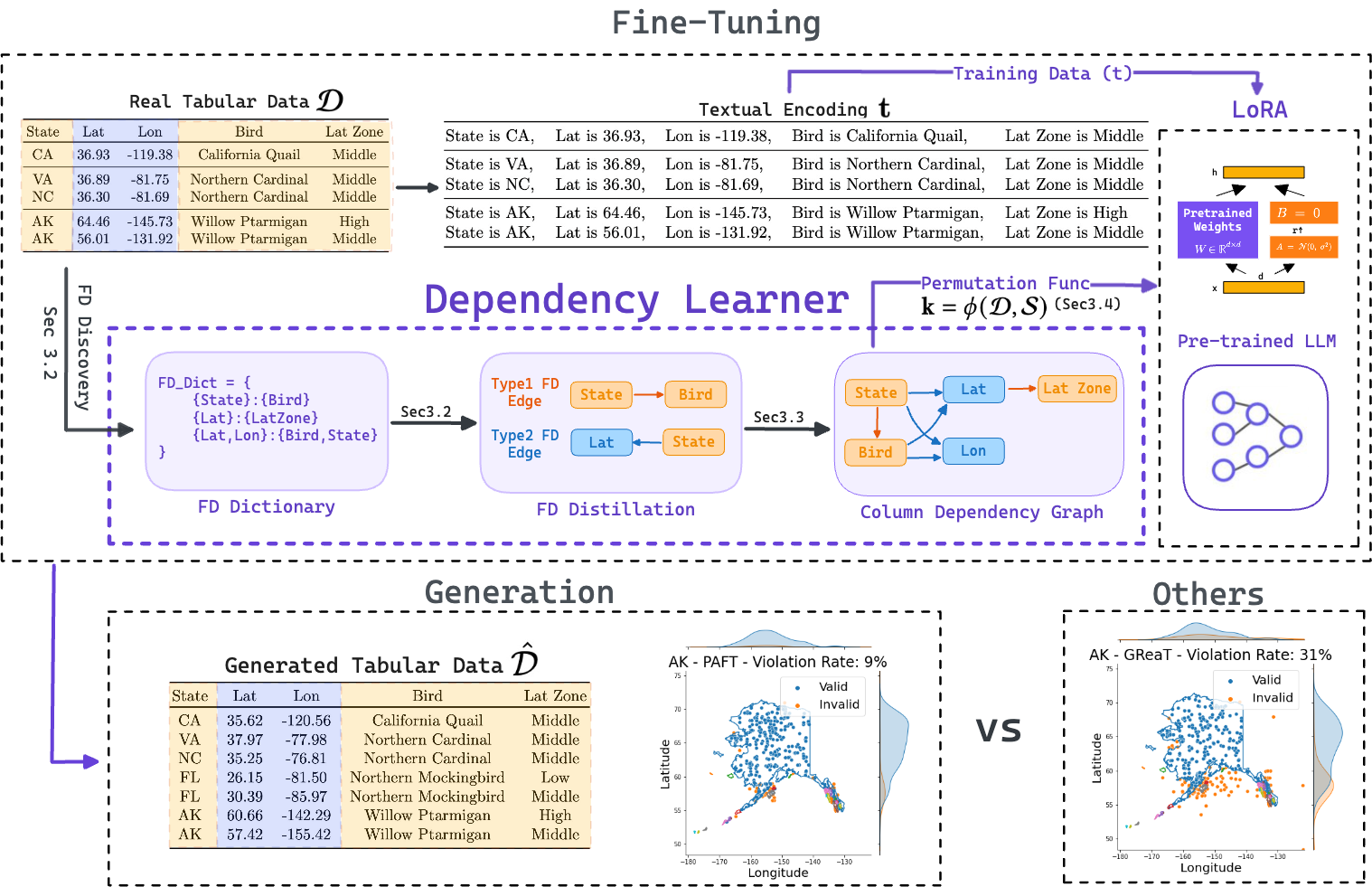}
    \caption{Overview of the proposed Permutation-Aided Fine-tuning (\ourmethod) approach.}
    \label{fig:sec1_overview}
\end{figure*}

Admittedly, the sure-fire way to check if a dataset has been faithfully modeled is to see if the joint
distribution of its features is captured accurately. 
Most current tests of synthetic data
generation quality focus on
fidelity to single-column
distributions, or to multi-column distributions.
For multiple variables, measures such as machine learning efficiency (MLE)~\cite{borisov2022language,xu2019modeling,zhang2023mixed,dinh2022lift} are frequently used to construct classifiers or regressors.

A key lesson 
from probabilistic graphical model research~\cite{koller2009probabilistic} is that
factorizing joint distributions into products of conditionals (e.g., Bayesian networks) dramatically helps reduce the number of parameters necessary to capture the underlying data characteristics. 
A similar lesson
from
database research~\cite{10.5555/1450931}
is that modeling functional
dependencies (FDs) in data helps reduce redundancy in modeling and storage.
These lessons, i.e., that
{\bf order matters},
continue to apply in the LLM era but are not internalized in our prompt ordering, fine-tuning, or evaluation
methodologies. 
Other researchers have
noted the
importance of
ordering in LLMs~\cite{chen2024premise,prystawski2024think} but
this lesson has 
not been leveraged to
improve
synthetic table generation
by LLMs.
In the absence of leveraging 
good feature orders, existing approaches either focus on `one order', `no order', or `all orders'.
All of these approaches 
severely limit the
quality of generated synthetic data.

As an example,
Table~\ref{tab:sec1_motivative_example} presents a case study
on using models to
generate synthetic data of locations in various states of the USA. The data is
of the form (state, latitude, longitude) 
where the attributes adhere to the FD: $\{\text{latitude}, \text{longitude}\} \rightarrow \text{state}$. 
Existing
methods can generate satisfactory univariate distributions (as shown in the border of the top row plots)
but fail
to capture the joint distribution (center of the top row plots) and conditional distributions across subcategories (bottom row plots).

In summary, feature ordering can be both a nuisance and a gift. It is a nuisance because it demands additional constraints to be modeled. It can be a gift because it suggests ways to sequentially generate data even by autoregressive LLMs. Our proposed approach (\ourmethod{}) aims to achieve an optimal permutation order for fine-tuning LLMs.

Figure~\ref{fig:sec1_overview} shows an overview of the proposed permutation aided fine-tuning approach (\ourmethod). A typical workflow is 1) Textual Encoding (Section~\ref{sec:tabdatagen_llm}) 2) Functional Dependency (FD) discovery (Section~\ref{sec:fd_discovery}) 3) FD Distillation ( Section~\ref{sec:fd_discovery}) and 4) Feature Order Permutation Optimization ( Section~\ref{sec:graph_search}). Our fine-tuning and sampling strategy is explained in Section~\ref{sec:fine_tuning}.

\section{\ourmethod: Permutation-Aided Fine-Tuning}

\textbf{Problem Setup}. Let $\mathcal{D}$ represent a table with $n$ rows (i.e., records) and $m$ columns (i.e., attributes a.k.a schema). Let each record be represented by vector $\mathbf{x}_{i}$ and further let $x_{ij}$ represent the element value of the $j^{th}$ attribute of record $\mathbf{x}_i$.
Hence each row $\mathbf{x}_i \in \mathcal{D}$ represents an individual record and each column $\mathbf{x}_{(:,j)}\sim \mathcal{X}_j$ can be considered sampled from a random variable $\mathcal{X}_j$ that governs the distribution of attribute $j$. Finally, let $i \in [1..n]$ and $j \in [1..m]$. Realistically, tabular data $\mathcal{D}$ is frequently a mixture of categorical and continuous attributes, hence each $\mathcal{X}_j$ can be a categorical or continuous random variable. If $\mathcal{A} = \{\mathcal{X}_1,\mathcal{X}_2,\dots,\mathcal{X}_m\}$ represents the collection of random variables, then the table generation process aims to sample from a joint distribution $\mathbb{P}(\mathcal{A}) = \mathbb{P}(\mathcal{X}_1,\mathcal{X}_2,\dots,\mathcal{X}_m)$. This joint distribution is usually a complex, high-dimensional distribution and, most importantly, unknown. 
The goal of learning an effective tabular data generator $p_\theta(\cdot)$ is to enable $p_\theta(\cdot)$ to learn a \emph{faithful} approximation $\mathbb{P}(\mathcal{A} | \mathcal{D})$ of the data generation process distribution $\mathbb{P}(\mathcal{A})$ using the data sample $\mathcal{D}$ such that $\mathbb{P}(\mathcal{A}|\mathcal{D}) \approx \mathbb{P}(\mathcal{A})$. Once such an effective model $p_{\theta}(\mathcal{D})$ is trained, it can be employed to generate large volumes of seemingly \emph{realistic} synthetic data $\mathcal{\hat{D}}\sim \mathbb{P}(\mathcal{A} | \mathcal{D})$.

\subsection{Tabular Data Generation with LLMs}\label{sec:tabdatagen_llm}
While training $p_{\theta}(\cdot)$, it is usually assumed that all records $\mathbf{x}_i \in \mathcal{D}$ are independent. Generating new data samples $\hat{\mathbf{x}}_i \in \hat{\mathcal{D}}$ can be done in various ways (e.g., see~\cite{xu2018synthesizing,xu2019modeling,zhao2021ctab}) which aim to directly estimate the joint distribution $\mathbb{P}(\mathcal{A})$ or, as is done here in \ourmethod{}, where $\mathbb{P}(\mathcal{A})$ is estimated by an autoregressive LLM based generative process, as a product of multiple conditional densities governed by the input ordering. 

Autoregressive LLM models are pre-trained to maximize the likelihood of \emph{target} token $x_{ij} \in \mathcal{D}$, conditioned upon the autoregressive context $\mathbf{x}_{(i,1:j-1)} \in \mathcal{D}$ where $\mathcal{D}$ is the training corpus comprising a large amount of textual data (in the pre-training context). Eq.~\ref{eq:mle_llm} defines the general training criterion of LLM training using the self-supervised next-token prediction task with `w' denoting the context length. 

\begin{equation}
    \mathcal{L}(\theta;\mathcal{D}) = -\hspace{-1ex}\sum\limits_{\mathbf{x}_i\in \mathcal{D}}\sum\limits_{j=1}^{w} \mathrm{log}\,\mathbb{P}(x_i|\mathbf{x}_{(i,1:j-1)}).
    \label{eq:mle_llm}
\end{equation}

The generation of a single instance (i.e., database record) $\mathbf{x}_i \in \mathcal{D}$ is given by Eq.~\ref{eq:joint_distribution}:

\begin{equation}
\label{eq:joint_distribution}
    \mathbb{P}(\textbf{x}_{i}) = \mathbb{P}(x_{i,1},...,x_{i,m}) \simeq \prod_{j=1}^{m} \mathbb{P}(x_{i,j} | x_{i,1},..,x_{i,j-1}).
\end{equation}

Specifically, each database record is generated as a product of conditional distributions.

\textbf{Input Encoding.} To support
the processing
of our
records $\mathbf{x}_i \in \mathcal{D}$ 
by a pre-trained LLM,
we adopt the following encoding:

\begin{equation}
\begin{split}
    t_{i,j} = [c_j, \textrm`is\textrm', x_{i,j}, \textrm`,\textrm'], & i \in \{1, .., n\}, j \in \{1, .., m\},\\
    \mathbf{t}_i = [t_{i,1}, t_{i,2}, .. , t_{i,m}],  & i \in \{1, .., n\}.
\end{split}
\label{eq:textual_encoder}
\end{equation}
In Eq.~\ref{eq:textual_encoder}, $c_j$ represents the attribute name of the $j^{th}$ database column while $x_{i,j}$ represents the actual value of the $j^{th}$ column for the $i^{th}$ record.
Further, we can assume we have a mechanism to obtain a \textit{feature order permutation} $\mathbf{k}$ to govern the order of the attributes in $\mathbf{t}_i$, such that $\mathbf{t}_i(\mathbf{k}) = [t_{i,k_1}, t_{i,k_2},..,t_{i,k_m}]$ (where $i\in \{1,..,n\}, k_j \in \{1,..,m\}$.), represents the same record but with the attribute order governed by the permutation $\mathbf{k}$,
 This definition admits the random feature order as a special case in which $\mathbf{k}$ is a random permutation. 

Since we consider autoregressive LLM-based generative models, employing the chain rule to sequentially produce each column of a table record $\mathbf{t}_i$, we can view  each generation
step as  \emph{approximating} the joint distribution of the table columns as a product of conditional distributions (i.e., $ \mathbb{P}(t_{i,1},...,t_{i,m}) \simeq \prod_{j=1}^{m} \mathbb{P}(t_{i,j} | t_{i,1},..,t_{i,{j-1}})$). 
However, as the number of columns increases and the relationships between columns get more conditional, the likelihood of encountering training and sampling bias due to class imbalance also rises~\cite{xu2019modeling}. To minimize such adverse effects, \ul{we can consider injecting knowledge of the pre-existing functional relationships among columns, to govern the autoregressive generation process}. To infer such functional relationships, we leverage a learned dependency graph derived from functional dependency (FD) relations which enables us to effectively determine the appropriate training and sampling sequence. This, in turn, allows us to alleviate potential biases during training by establishing a \emph{generation curriculum} leading to improved estimation accuracy of the joint distribution $\mathbb{P}(\textbf{t}_{i})$ in auto-regressive prediction $ \mathbb{P}(t_{i,k_1},...,t_{i,{k_{m}}}) \simeq \prod_{j=1}^{m} \mathbb{P}(t_{i,k_j} | t_{i,k_1},..,t_{i,k_{j-1}})$, where the ordering $t_{i,k_1}\dots,t_{i,k_m}$ is obtained by a feature order permutation function $\mathbf{k} = \phi(\mathcal{D},\mathcal{S})$. We detail the requisite background and design of $\phi(\mathcal{D},\mathcal{S})$ in sections~\ref{sec:fd_discovery} and ~\ref{sec:graph_search}.

\subsection{Discovery and Distillation of Functional Dependencies (FD)}
\label{sec:fd_discovery}

A functional dependency (FD) is a relationship $R$ in schema $S$ that exists when a subset of attributes $ A \subset S$ uniquely determines another subset 
$B \subset S$
of attributes. We succinctly represent an FD as $R: A \rightarrow B$ which specifies that $B$ is functionally dependent on $A$.

\begin{definition}[Schema-Level FD]
    \label{def:table_fd}
    With $A$, $B$ being two disjoint subsets of the schema (columns) of table $\mathcal{D}$, a schema-level FD $F$ associated with $\mathcal{D}$ has the form:
    $F: A \rightarrow B$.
\end{definition}

We leverage FD discovery techniques to govern the order of the autoregressive data generation process in \ourmethod{}. A large body of
research from the database literature on FD discovery~\cite{papenbrock2015functional,zhang2020statistical,papenbrock2016hybrid} can be leveraged in \ourmethod{},
including methods that account for noisy FDs~\cite{zhang2020statistical}. In this work, we focus on leveraging
schema-level FDs
discovered using
a state-of-the-art FD discovery algorithm~\cite{papenbrock2016hybrid} to govern the autoregressive data generation process.

\par\noindent
\textbf{FD Distillation}. The result of traditional FD discovery, yields complex (i.e., multi-attribute) functional dependencies between columns which are ambiguous to resolve in an autoregressive generation setting. Hence we undertake an intermediate \emph{FD distillation} step to simplify multi-attribute functional dependencies into multiple single attribute FDs as detailed below.

\begin{figure}[htb]
     \centering
     \begin{subfigure}[b]{0.8\linewidth}
         \centering
         \includegraphics[width=\linewidth]{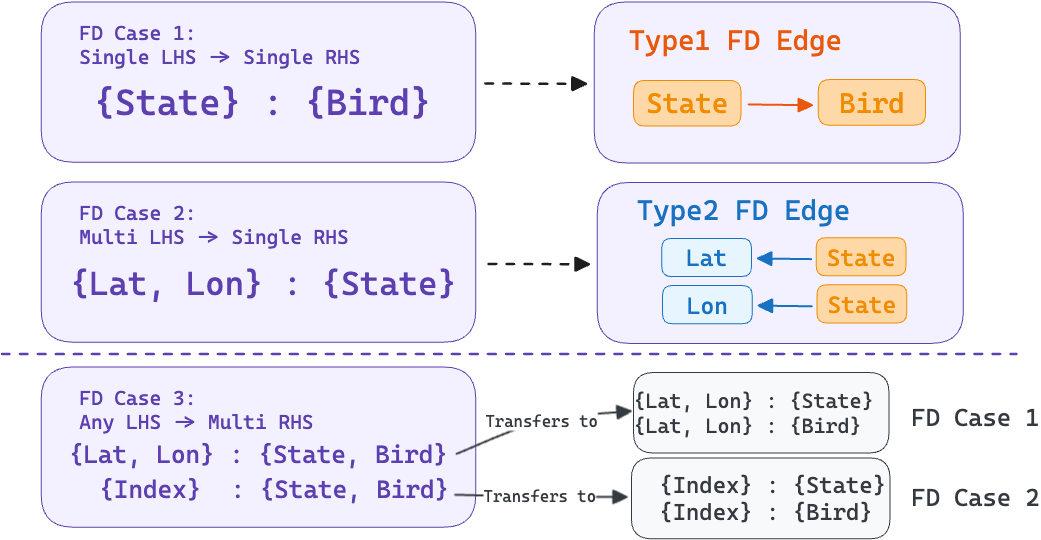}
         \caption{There are two types of column dependency edges for three types of functional dependencies (FDs), which are distinguished by the left-hand side (LHS) and right-hand side (RHS) in the FD.}
         \label{fig:edge_concept}
     \end{subfigure}
 \hfill
     \begin{subfigure}[b]{0.8\linewidth}
         \centering
         \includegraphics[width=\linewidth]{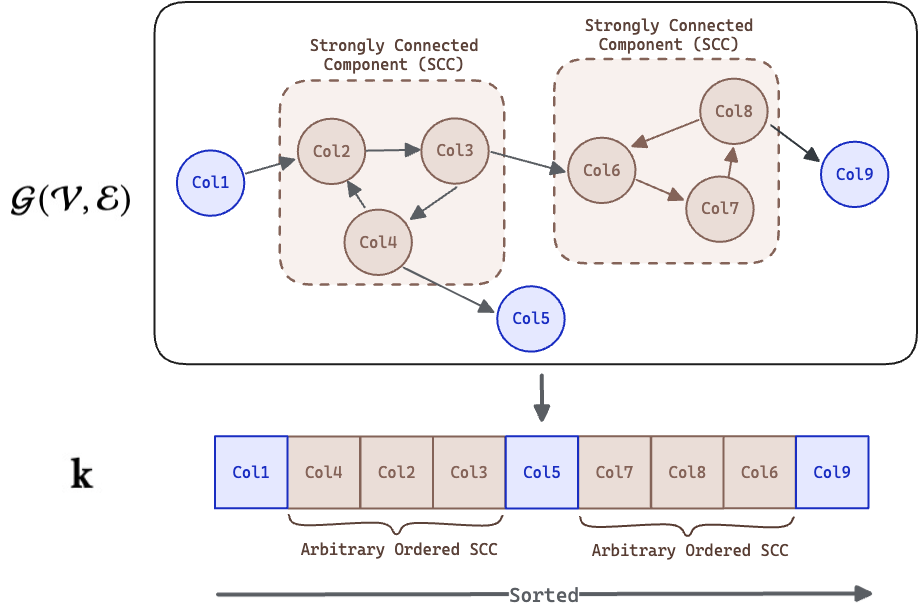}
         \caption{DAG for column functional dependency derived by expanding SCC super nodes and retrieving a fully flattened, ordered structure.}
         \label{fig:edge_scc}
     \end{subfigure}
    \caption{FD-Distillation and Dependency Graph Sorting for automatically extracting order permutations from tables.}
        \label{fig:two_method_graphs}
\vspace{-10pt}
\end{figure}

We first construct a dependency graph model $G(\mathcal{V}, \mathcal{E})$ where $\mathcal{V}$ represents the set of vertices with each $v \in \mathcal{V}$ representing an attribute in $\mathcal{S}$ and $e_{ij} \in \mathcal{E}$ representing an edge relation from attribute $v_i$ to attribute $v_j$. Further, we consider two types of edges in $\mathcal{E}$, specifically each $e_{ij}$ may be a \emph{type-1} edge or a \emph{type-2} edge (defined next).
The two edge types (i.e., \emph{type-1}, \emph{type-2}) in $\mathcal{E}$ are derived from three classes
of FDs,
as shown in
Fig.~\ref{fig:edge_concept}.
Subsequently, we proceed to examine each individual case: 1) Single attribute left-hand side (LHS) and single attribute right-hand side (RHS) 2) Multi-attribute LHS and single-attribute RHS. 3) single or multi-attribute LHS and multi-attribute RHS. We shall use the example table in Fig. ~\ref{fig:sec1_overview} to define each FD case.

[\textbf{Type-1 Edge}]. Let us consider an example of FD Case 1, wherein the column \emph{State} 
functionally
determines  column \emph{Bird}. In such a case, we enforce that the value for the attribute \emph{State} be generated prior to the value for the attribute \emph{Bird}. Accordingly,
a forward directed edge from \emph{State} to \emph{Bird} is created in the column dependency graph $\mathcal{G}$. We term such forward directed edges as type-1 edges in $\mathcal{G}$.
[\textbf{Type-2 Edge}]. The other type of edge in $\mathcal{G}$, arises when we encounter an FD with a multi-attribute LHS and a single attribute RHS (i.e., FD Case 2). As per FD Case 2, the values of multiple columns in the LHS would collectively decide the value of the column on the RHS. As an example in Fig.~\ref{fig:edge_concept}, the tuple of columns \emph{Latitude, Longitude} functionally
determines \emph{State}.
For such FDs, two backward edges are added in the column dependency graph $\mathcal{G}$, connecting \emph{State} to both \emph{Latitude} and \emph{Longitude}. We term such backward directed edges as type-2 edges in $\mathcal{G}$.

FD Case 3 relationships are ones where the RHS has multiple attributes and the LHS could have single or multiple-attributes. Such relationships do not directly result in an edge in our dependency graph $\mathcal{G}$. Instead, as classically done in FD literature~\cite{papenbrock2016hybrid}, we subject such FD relationships to an intermediate decomposition step. Specifically, the multi-attribute RHS of FD Case 3 relationships is decomposed into multiple single-attribute RHS dependencies each comprising the original LHS. Further, in each of these new decomposed relationships, if the LHS is single-attribute, it is treated as a FD \emph{Case 1} relationship (i.e., a directed edge from $LHS$ to $RHS$ is added to $\mathcal{G}$), else it is handled as an FD \emph{Case 2} relationship, wherein for each attribute in the multi-attribute LHS, a \emph{backward} dependency edge from the single-attribute RHS is added to the dependency graph $\mathcal{G}$.
Therefore, for every column in the right-hand side (RHS), we employ either \emph{type-1} or \emph{type-2} edge construction, depending on the value of its left-hand side (LHS).

\subsection{Putting It All Together}
\label{sec:graph_search}
Until this point, our construction of the graph $\mathcal{G}$ has only been limited to considering pair-wise relationships between columns in $\mathcal{D}$. Graph properties like functional dependency transitivity, require us to obtain a total ordering on the nodes $v \in \mathcal{V}$ of $\mathcal{G}(\mathcal{V},\mathcal{E})$ that is deterministic in nature for effective auto-regressive LLM training. This implies that in order to obtain a feature order permutation ($\mathbf{k}$) from the derived functional dependency relationships (Sec.~\ref{sec:fd_discovery}), a computation must be performed on the entire dependency graph $G(\mathcal{V}, \mathcal{E})$.

We define this task of obtaining a total feature order $\mathbf{k}$ from $\mathcal{G}(\mathcal{V},\mathcal{E})$ as an optimization step which seeks to produce $\mathbf{k}$ while minimizing the number of violated relationships in $\mathcal{G}(\mathcal{V},\mathcal{E})$.

Consider the trivial case of having an empty FD graph $\mathcal{G}$ i.e., $|\mathcal{E}| = \emptyset$.  If the columns of a table are not functionally dependent on each other, then the order of generation is not important and $\mathbf{k}$ can be some arbitrary permutation of the columns in $\mathcal{S}$.
For all other cases, our total feature ordering algorithm operates in three phases, as shown in Fig.~\ref{fig:edge_scc}.
[\textbf{Phase 1: Condensation}]. It is apparent that if a graph is not a directed acyclic graph (DAG), there is no optimal solution to the total feature ordering problem. In other words, there must be FDs that cannot be satisfied in the resulting total order permutation $\mathbf{k}$. 
In such cases, we compute the strongly connected components (SCC), and 
condense them into super nodes, thus transforming the original graph into a DAG.
[\textbf{Phase 2: Ordering}]. An application of a topological sort onto the DAG from Phase 1 will result in a  total feature ordering with all SCCs in $\mathcal{G}$ compressed into super nodes.
[\textbf{Phase 3: Expansion of SCC}]. Once the topological sort is conducted in Phase 2, we finally expand the SCC super nodes (via. arbitrary ordering) such that although the intra-SCC ordering of the nodes within the SCC is arbitrary, their ordering relative to non-SCC nodes is maintained.

\subsection{Synthetic Data Generation using \ourmethod{}}
\label{sec:fine_tuning}

After the optimized feature order permutation is obtained, we fine-tune the LLM with the textually encoded table record $\mathbf{t}_i$ 
such that the auto-regressive generation process is governed by the optimal feature order permutation $\mathbf{k} = \phi(\mathcal{D},S)$. Specifically, we generate the table governed by order 
$\mathbf{k}$ as defined in Eq.~\ref{eq:joint_distribution_with_permutation}.

\begin{equation}
\label{eq:joint_distribution_with_permutation}
    \mathbb{P}(\textbf{t}_{i}) = \mathbb{P}(t_{i,k_1},...,t_{i,{k_{j}}}) \simeq \prod_{j=1}^{m} \mathbb{P}(t_{i,k_j} | t_{i,k_1},..,t_{i,k_{j-1}}).
\end{equation}

We employ 
the Low-Rank Adaptation (LoRA) fine-tuning strategy~\cite{LoRA}.  To generate
synthetic rows, 
we first sample the initial token $p(t_{i,k_1})$ from the marginal distribution of variable $k_1$ in actual training data,
and then use
Eq.~\ref{eq:joint_distribution_with_permutation} to sequentially sample subsequent tokens $p(t_{i,k_j})$, where $j \in {2,...,m}$.

\section{Experimental Evaluation}
We conduct an exhaustive empirical
evaluation of \ourmethod to assess
its ability to reproduce
realistic data distributions,
superiority over other competing
approaches, and most importantly
substitutability of data
generated by \ourmethod in the context of a larger ML pipeline. More specifically, the questions we seek to answer are:

\begin{enumerate}
\item Does PAFT-generated synthetic data
accurately capture conditional distributions
within categories? (Sec~\ref{sec:conditional_distribution})
\item Does PAFT generate data respect
the consistency of intrinsic data
characteristics? (Sec~\ref{sec:eval_domain_knowledge})
\item Does the synthetic data generated by PAFT pass the \emph{sniff} test?
(Sec~\ref{sec:eval_discrimnator})
\item Can data generated by PAFT replace real data in downstream ML
model training? 
(Sec~\ref{sec:eval_mle})
\item  Do the data sets generated by \ourmethod adhere to real distributions and possess mode diversity?
(Sec~\ref{sec:eval_diversity})
\item Do newer generations of LLMs obviate the need for \ourmethod{}?
(Sec~\ref{sec:discussion_gpt4})
\end{enumerate}

\textbf{Datasets.} We evaluate the effectiveness of \ourmethod through experiments on six real datasets commonly used in synthetic table generation studies such as GReaT~\cite{borisov2022language} CTGAN~\cite{xu2019modeling}. These are Beijing~\cite{beijing}, US-locations~\cite{arcgisStatesShapefile}, California Housing~\cite{california}, Adult Income~\cite{adult}, Seattle~\cite{seattle}, and Travel~\cite{travel}. Separately, we also generate a set of four simulated datasets for class-mixture distributions.

\textbf{Baselines.} For benchmarking, we organize baselines that utilize current deep learning approaches for synthetic data generation (CTGAN~\cite{xu2019modeling}, CopulaGAN~\cite{xu2019modeling}, TabSyn~\cite{zhang2023mixed}) and the most advanced synthetic table generator with LLM fine-tuning GReaT~\cite{borisov2022language}. 
To guarantee an equitable comparison, we employ the Distill-GReaT model for both LLM techniques in all tests.

\textbf{Reproducibility.} Each baseline (CTGAN, CopulaGAN, TabSyn, GReaT) adheres to the recommended hyperparameters and utilizes officially released API tools: Synthetic Data Vault~\cite{SDV} and GReaT~\cite{borisov2022language}. For a fair comparison of GReaT and PAFT, the LoRA fine-tuning parameters are set the same as: Lora attention dimension $r=16$, alpha parameter for Lora scaling $lora\_alpha=32$, the names of the modules to apply the adapter to $target\_modules=c\_attn$, the dropout probability for Lora layers $lora\_dropout=0.05$, $bias=none$.

\textbf{Parameters for MLE and Discriminator Models.}
We utilize neural network, linear/logistic regression, and random forest models from the Scikit-Learn package for the ML efficiency and discriminator experiments. 
 Every result is evaluated through the 5-fold cross-validation process.
 
\textbf{Low-order statistics}~\cite{zhang2023mixed} of column-wise data distribution and pair-column correlation are calculated with the SDV library
\footnote{\url{https://docs.sdv.dev/sdmetrics/reports/quality-report}\label{foot:sdm}}.

\subsection{RQ1: Does PAFT-generated synthetic data accurately capture conditional distributions within categories?}
\label{sec:conditional_distribution}

\begin{figure}[tpb!]
\centering
\includegraphics[width=.87\linewidth]{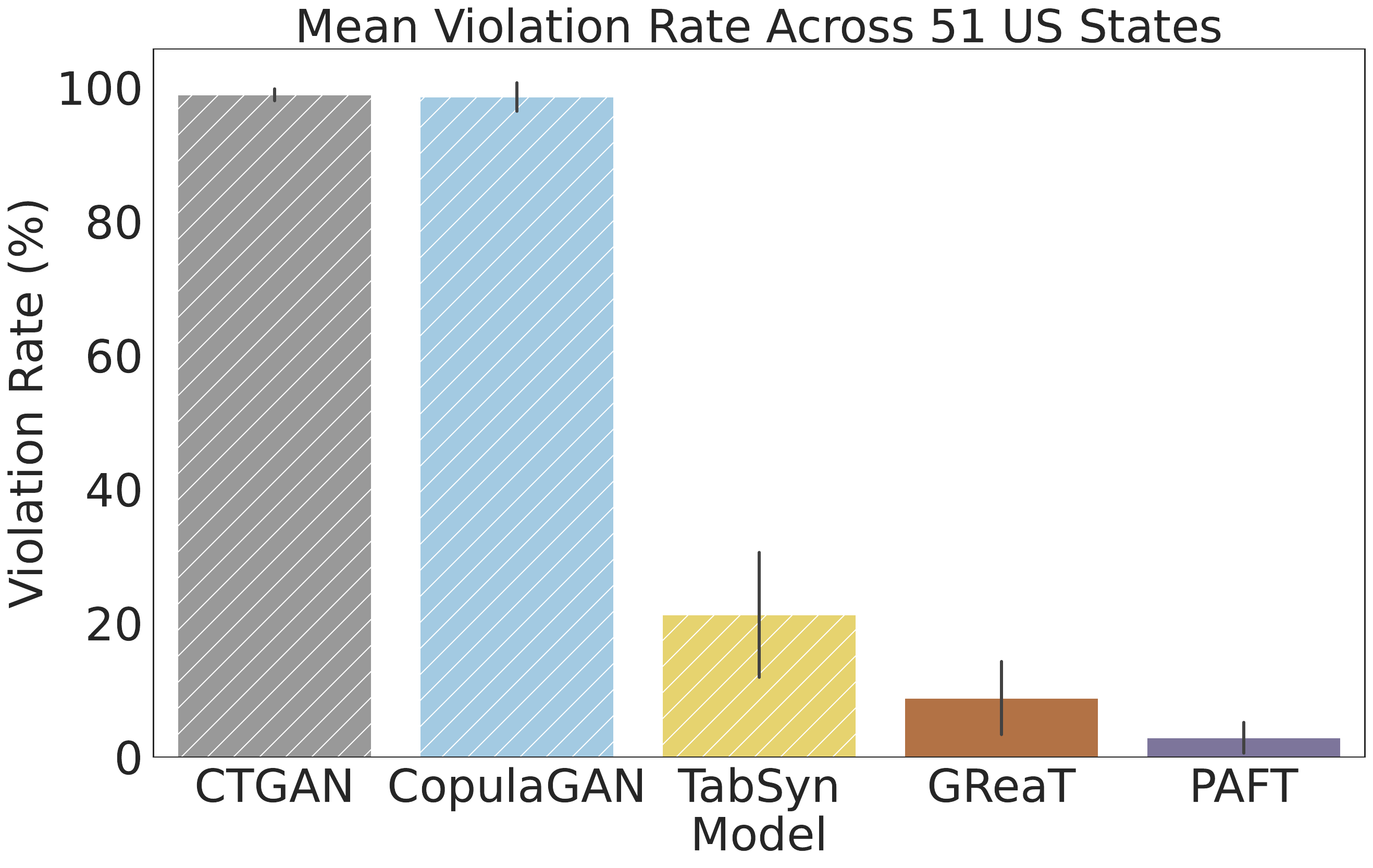}
\caption{For a composite dataset, US-locations, this comparison examines state-specific violation rates across different synthetic data generation approaches. The error bars represent standard deviation. The states on the x-axis are ordered by decreasing violation rates. PAFT significantly reduces state-specific violations in the composite dataset.}
\label{fig:violation_rate_by_state_bar}
\end{figure}

\begin{table}[htpb]
    \centering
    \caption{Violation rates across different categories in the same dataset highlight the effectiveness of PAFT in addressing conditional distribution challenges in mixed-category datasets. Notably, even though baseline models may perform well in the MLE task or distribution evaluation, they often fail in practical boundary checks, such as functional dependencies (FDs). \textbf{We can see that states with the lowest violation rates are the easiest to model, i.e. rectangular.}}
    \label{tab:sec4_state_geo_figure}
    \setlength{\tabcolsep}{1mm}
    \begin{tabular}{c|ccccc}
        \toprule
        \multicolumn{6}{c}{\makecell{States with the \textbf{\underline{Highest}} Violation Rates ($\downarrow$)\\(Sorted by LLM baseline: GReaT)}} \\
        \midrule
        Category (State) & MO & AK & KY & FL & WV \\
        & \includegraphics[width=.1\linewidth,height=1cm]{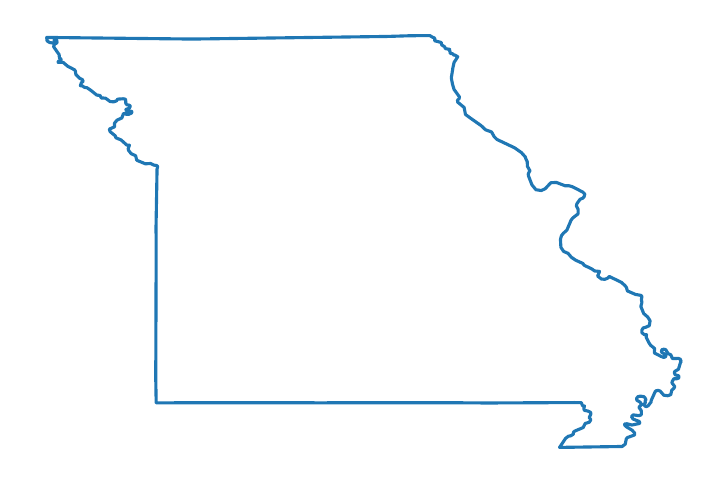}
        & \includegraphics[width=.1\linewidth,height=1cm]{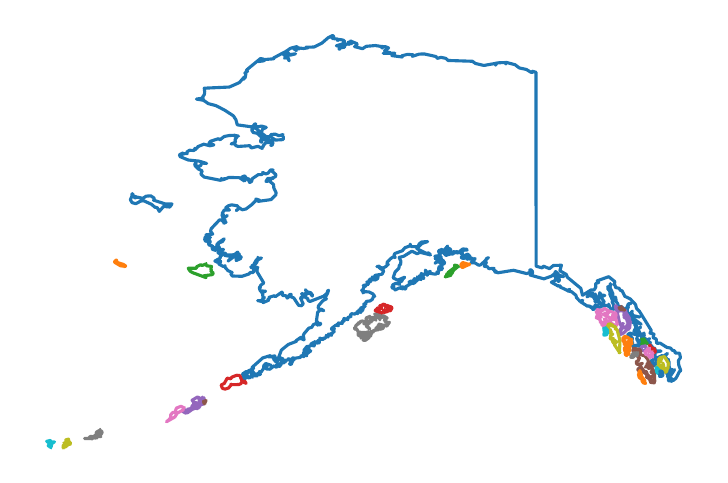}
        & \includegraphics[width=.1\linewidth,height=1cm]{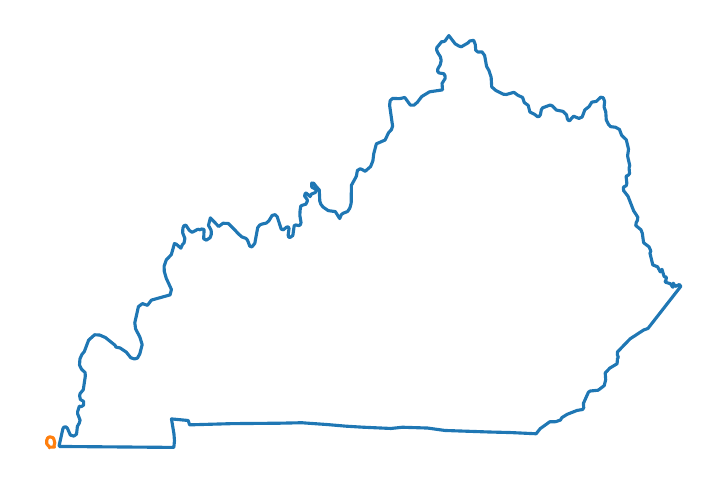}
        & \includegraphics[width=.1\linewidth,height=1cm]{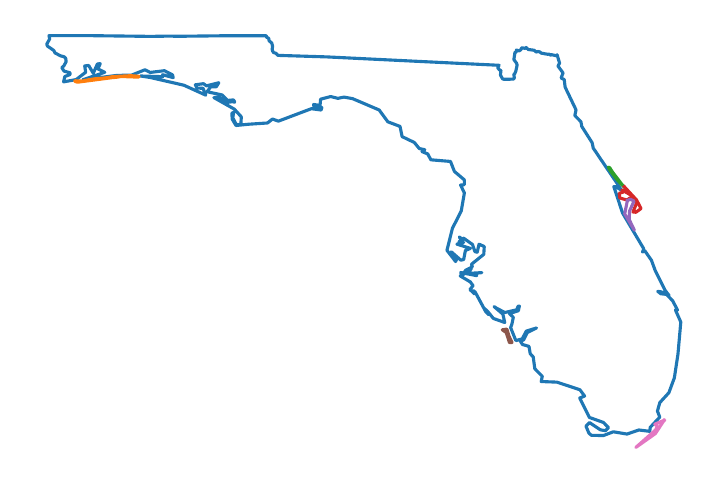}
        & \includegraphics[width=.1\linewidth,height=1cm]{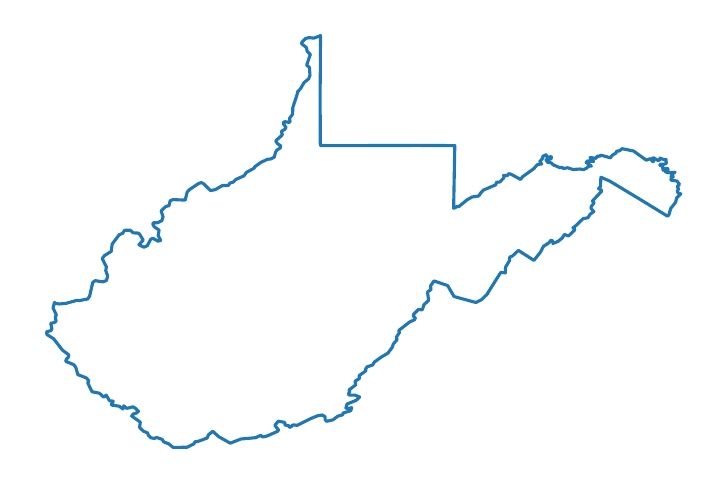}
        \\
        \midrule
        Real Data & 0\% & 0\%&0\%&0\%& 0\%\\
        \midrule
        CTGAN & 98.0\% & 99.3\%&97.4\%&99.9\%& 100\%\\
        CopulaGAN & 99.3\% & 84.6\%&99.3\%&97.8\%& 100\%\\
        TabSyn & 10.5\% & 17.2\%&22.9\%&25.1\%& 28.1\%\\
        GReaT & 17.1\% &18.4\%&20.6\%&21.9\%&22.3\%\\
        \midrule
        PAFT &\textbf{1.9\%} &\textbf{5.1\%}&\textbf{3.4\%}&\textbf{3.4\%}&\textbf{3.4\%}\\
        \bottomrule
        \bottomrule
        \multicolumn{6}{c}{\makecell{States with the \textbf{\underline{Lowest}} Violation Rates ($\downarrow$)\\(Sorted by LLM baseline: GReaT)}} \\
        \midrule
        Category (State) & CO & KS & NM & SD & UT \\
        & \includegraphics[width=.1\linewidth,height=1cm]{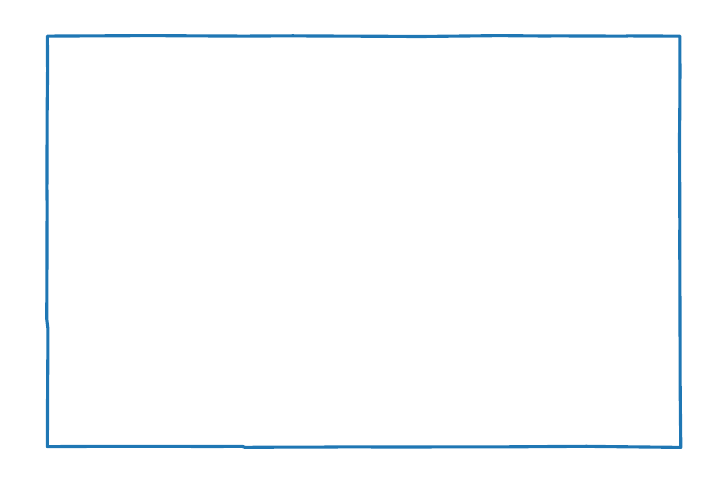}
        & \includegraphics[width=.1\linewidth,height=1cm]{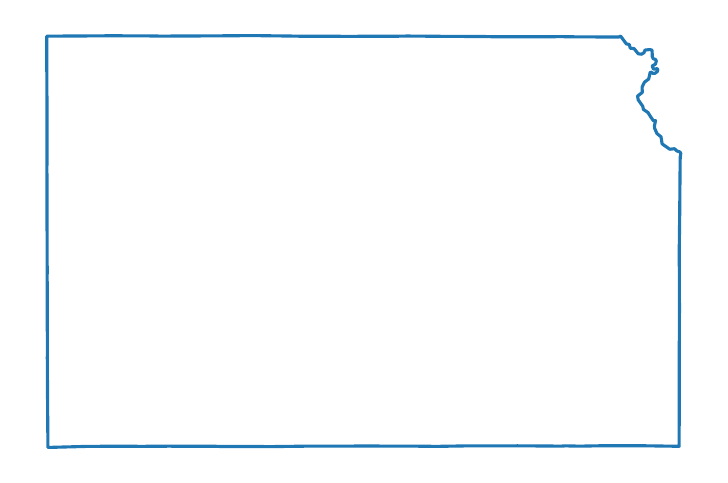}
        & \includegraphics[width=.1\linewidth,height=1cm]{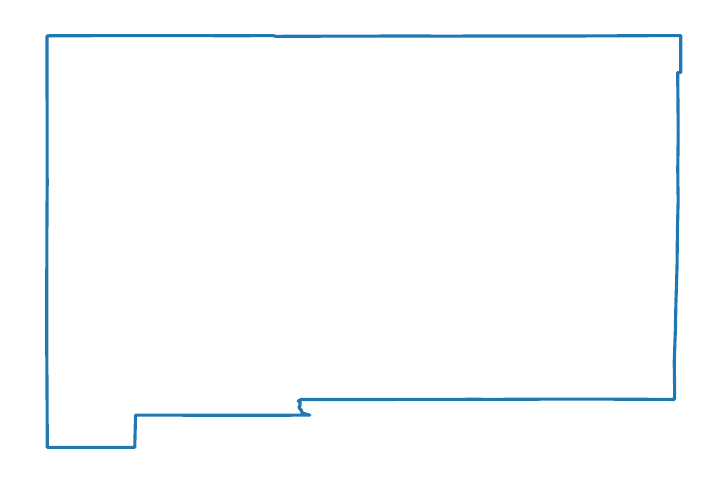}
        & \includegraphics[width=.1\linewidth,height=1cm]{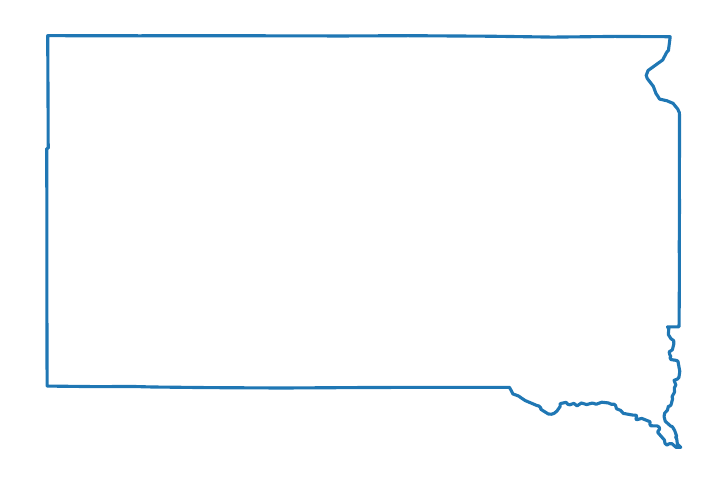}
        & \includegraphics[width=.1\linewidth,height=1cm]{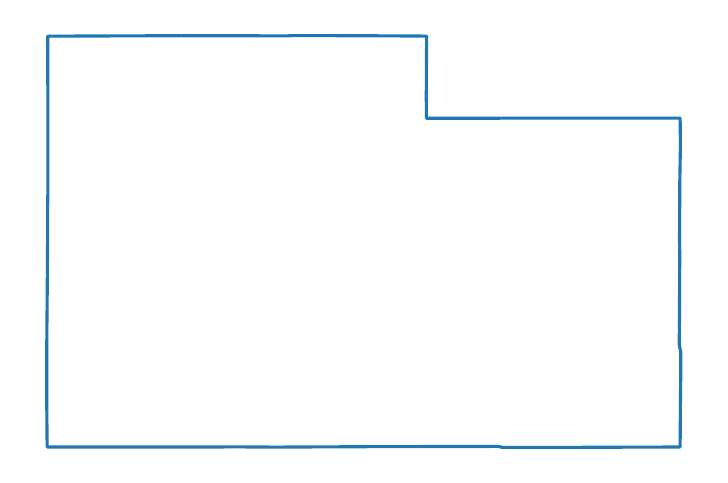}
        \\
        \midrule
        Real Data & 0\% & 0\%&0\%&0\%& 0\%\\
        \midrule
        CTGAN & 96.5\% & 97.0\%&99.5\%&97.5\%& 99.3\%\\
        CopulaGAN & 98.3\% & 98.8\%&98.4\%&98.3\%& 98.4\%\\
        TabSyn & 11.3\% & 18.4\%&6.2\%&17.3\%& 16.6\%\\
        GReaT & 0.5\% &0.9\%&1.2\%&1.5\%&2.1\%\\
        \midrule
        PAFT &\textbf{0.0\%} & \textbf{0.3\%}&\textbf{0.9\%}&\textbf{0.9\%}&\textbf{0.0\%}\\
        \bottomrule
    \end{tabular}
\end{table}

Capturing and generating the diversity in a multi-class setting (composite dataset) has been shown to be a challenging practical problem~\cite{margeloiutabebm}.
In addition to Table~\ref{tab:sec1_motivative_example}, Fig. \ref{fig:violation_rate_by_state_bar} and Table \ref{tab:sec4_state_geo_figure} further illustrate the widespread nature of this issue within the same composite dataset, where the GAN-generated results exhibit an almost 100\% violation rate. The term ‘composite dataset’ refers to a dataset in which the distributions across different subcategories show significant variances.
The LLM model GReaT, which employs random order permutation, shows a significant improvement in maintaining conditional distributions. Nevertheless, its performance remains inconsistent due to unevenness within categories, resulting in violation rates ranging from 0.5\% to 22.3\%.
In contrast, \ourmethod consistently controls deviations from the real facts of conditional distributions, maintaining them within a range of 0\% to 5\%. This reliability holds even in challenging subcategory cases where baseline methods underperform.

\subsection{RQ2: Does PAFT generate data respecting the consistency of intrinsic data characteristics?}
\label{sec:eval_domain_knowledge}

\begin{table}[tb]
    \centering
    \caption{Datasets have intrinsic characteristics like functional dependencies, range restrictions, and other domain knowledge. Results are averaged over five random runs.}
    \label{tab:knowledge-check}
    \setlength{\tabcolsep}{0.3mm}
    \begin{tabular}{llccccc}
        \toprule
        \multirow{2}{*}{\makecell{\textbf{Intrinsic Fact}\\(\textbf{Dataset})}} &  \multirow{2}{*}{} & 
        \multicolumn{5}{c}{\textbf{Fact Violation Rate ($\downarrow$)}} \\
        \cmidrule{3-7}
         &  & CTGAN & Cop.GAN & TabSyn & GReaT & \ourmethod \\
        \midrule
        \makecell{Lat-long$\rightarrow$State\\(US-locations)} & & 99.2\% & 98.5\% & 21.5\% & 8.2\% & \bf 2.9\%\\
        \midrule
        \makecell{Lat-long $\rightarrow$ CA\\(California)} &  & 47.6\%& 99.9\% & 8.8\% & 5.4\% & \bf 1.3\%\\
        \midrule
        \makecell{Med. house price \\$\rightarrow$ [1.4$e^5$, 5$e^5$] \\(California)} & & 1.5\% & 0.01\% & \bf 0.0\% & \bf 0.0\% & \bf 0.0\%  \\
        \midrule
        \makecell{education$\rightarrow$\\ education-num\\(Adult)} &   & 83.9\% & 19.1\% & 1.4\% & 1.2\% & \textbf{0.5\%} \\
        \midrule
        \makecell{Zipcode$\rightarrow$Seattle\\(Seattle)} &  & \bf  0.0\% & 99.9\% & \bf 0.0\% & \bf 0.0\% & \bf 0.0\% \\
        \bottomrule
    \end{tabular}
\end{table}

In addition to the dissimilar sub-category challenge, we also investigate whether unsatisfactory conditional distributions exist in general across various datasets, and whether the PAFT method can address these issues.
In line with this, we conducted rule checks that were derived from real-world scenarios and subsequently evaluated the generated data from all models. Table~\ref{tab:knowledge-check} displays the violation rate in the generated data. From the table, we observe that \ourmethod adheres to data's characteristics more faithfully (i.e., significantly fewer rule violations) than baseline methods, by learning together with functional dependencies.

\subsection{RQ3: Does the synthetic data generated by PAFT pass the \emph{sniff} test?}
\label{sec:eval_discrimnator}

Similarly to the analysis conducted in recent work~\cite{borisov2022language} (GReaT), we employ the random forest (RF) algorithm to train discriminators to distinguish real data (labeled True) and synthetically generated data (labeled False). Subsequently, we test performance on an unseen set (consisting of 50\% synthetically generated data and 50\% real data). In this experiment, scores represent the percentage of correctly classified entities. In this case, an ideal accuracy score would be close to 50\%, which means the discrimniator fails to distinguish between real and synthesized data. The scores are shown in Table~\ref{tab:Discriminator} and indicate that the data generated by \ourmethod are most indistinguishable from real data, even by powerful discriminative models.

\begin{table}[htb!]
    \centering
    \caption{Privacy Discriminator Performance. The scores stand for the accuracy for detecting real or fake data, where the ML models are trained using 50\% real data and 50\% random data. An ideal accuracy score is 50, indicating the model cannot distinguish between real and synthesized data. The best results are marked in \textbf{bold}, the second-best results are \underline{underlined}. Results are averaged over five random runs.}
    \label{tab:Discriminator}
    \setlength{\tabcolsep}{0.5mm}
    \begin{tabular}{p{2cm}|c|c|c|c|c}
        \toprule
        \multicolumn{6}{c}{\makecell{\textbf{Data Sniff Test} - ML Discriminator Accuracy\\(Values closest to 50\% are best.)~}} \\
        \midrule
        
        \textbf{Method} & CTGAN & Co.GAN & TabSyn & GReaT & \textbf{\ourmethod} \\
        \midrule
        Beijing & 99.16\% & 98.69\% & \underline{50.97\%} & 51.1\% & \bf 50.09\% \\
        \midrule
        US-locations & 99.94\% & 97.74\% & 51.97\% & \underline{50.47\%} & \bf 50.01\% \\
        \midrule
        California & 98.35\% & 86.64\% & \underline{50.64\%} & 53.74\% & \bf 49.89\% \\
        \midrule
        Adult & 94.43\% & 59.82\% & 51.64\% & \bf 51.12\% & \underline{48.75\%} \\
        \midrule
        Seattle & 87.61\% & 85.7\% & \bf 50.12\% & 68.27\% & \underline{47.21\%} \\
        \midrule
        Travel & 77.96\% & 74.14\% & \bf 50.66\% & 62.49\% & \underline{48.18\%} \\
        \bottomrule
    \end{tabular}
\end{table}

\begin{table}[htb!]
    \centering
    \caption{MLE performance. 
    MLE performance: For datasets with regression tasks (marked *), performance is evaluated using MAPE (where lower scores are better). For datasets with classification tasks, accuracy is used (where higher scores are better).
    The best results are marked in \textbf{bold} and the second-best results are \underline{underlined}.
    Results are averaged over five random runs}. 
    \setlength{\tabcolsep}{0.1mm}
    \begin{tabular}{p{1cm}l|c|ccc|cc}
        \toprule
        \multicolumn{2}{l|}{\textbf{Regression}}& 
        \multicolumn{6}{c}{\textbf{MAPE ($\downarrow$)~} Over Different Methods} \\
        \midrule
        \textbf{Dataset(*)} & & Orig. &  CTGAN & Co.GAN & TabSyn & GReaT & \textbf{\ourmethod}\\
        \midrule
        \multirow{3}{*}{Beijing} & \textbf{RF} & 0.41\% & 2.49\% & 2.15\% & 0.7\% & \underline{0.57\%} & \bf 0.52\% \\
        & \textbf{LR} & 1.37\% & 2.23\% & 1.55\% & \underline{1.25\%} & \bf 0.97\% & 1.34\% \\ 
        & \textbf{NN} & 0.99\% & 2.44\% & 2.83\% & \underline{1.01\%} & 1.16\% & \bf 0.95\% \\
        \midrule
        \multirow{3}{*}{Calif.} & \textbf{RF} & 0.18\% & 0.65\% & 0.39\% & \underline{0.22\%} & 0.25\% & \bf 0.20\% \\
        & \textbf{LR} & 0.30\% & 0.54\% & 0.5\% & \underline{0.30\%} & \bf 0.29\% & 0.31\% \\ 
        & \textbf{NN} & 0.34\% & 0.53\% & 0.47\% & \underline{0.29\%} & 0.3\% & \bf 0.27\% \\
        \midrule
        \multirow{3}{*}{Seattle} & \textbf{RF} & 0.33\% & 0.76\% & 0.38\% & \underline{0.30\%} & 0.35\% & \bf 0.28\% \\
        & \textbf{LR} & 0.29\% & 0.74\% & 0.32\% & \bf 0.23\% & 0.33\% & \underline{0.29\%} \\ 
        & \textbf{NN} & 0.28\% & 0.71\% & 0.38\% & \underline{0.28\%} & 0.33\% & \bf 0.27\% \\
        \bottomrule
        \toprule
        \multicolumn{2}{l|}{\textbf{Classif.}} & 
        \multicolumn{6}{c}{\textbf{Accuracy ($\uparrow$)~} Over Different Methods} \\
        \midrule
        \textbf{Dataset} & & Orig. &  CTGAN & Co.GAN & TabSyn & GReaT & \textbf{\ourmethod}\\
        \midrule
        \multirow{3}{*}{\makecell[l]{US-loc.}} & \textbf{RF} & 99.95\% & 7.17\% & 45.33\% & \bf 99.99\% & 99.84\% & \underline{99.91\%} \\
        & \textbf{LR} & 46.1\% & 5.11\% & 31.08\% & 43.69\% & \underline{45.65\%} & \bf 49.41\% \\ 
        & \textbf{NN} & 99.85\% & 7.56\% & 53.34\% & \bf 99.64\% & 98.94\% & \underline{99.44\%} \\
        \midrule
        \multirow{3}{*}{Adult} & \textbf{RF} & 84.97\% & 71.15\% & 81.33\% & \underline{83.69\%} & \bf 83.89\% & 83.06\% \\
        & \textbf{LR} & 78.53\% & 75.68\% & 78.18\% & \bf 78.38\% & 76.1\% & \underline{77.24\%} \\ 
        & \textbf{NN} & 76.9\% & 75.69\% & 76.6\% & \underline{78.36\%} & 78.23\% & \bf 79.16\% \\
        \midrule
        \multirow{3}{*}{Travel} & \textbf{RF} & 88.95\% & 56.35\% & 67.18\% & \underline{84.09\%} & 79.78\% & \bf 85.19\% \\
        & \textbf{LR} & 82.87\% & 70.17\% & 79.56\% & \bf 83.31\% & 78.34\% & \underline{82.76\%} \\ 
        & \textbf{NN} & 81.77\% & 71.05\% & 79.56\% & \underline{81.88\%} & 80.77\% & \bf 83.20\% \\
        \bottomrule
    \end{tabular}
    \label{tab:MLE}
\end{table}

\subsection{RQ4: Can data generated by PAFT replace real data in downstream ML
model training?}
\label{sec:eval_mle}

We next assess the effectiveness of the generated (synthetic) data by comparing the performance of discriminative models trained on synthetic data versus real data for their target tasks. Models tested include random forests (RF), linear regression (LR), and multi-layer perceptron (NN).
As shown in Table~\ref{tab:MLE}, \ourmethod{} is best or second best in over 80\% of
(dataset, method) combinations.

  \subsection{RQ5: Do the data sets generated by \ourmethod adhere to real distributions and possess mode diversity?}
\label{sec:eval_diversity}

We also evaluate how closely the density and diversity of the true data distribution are matched by \ourmethod generated data using correlation metrics and density-based distance metrics. Specifically, we employ the Kolmogorov-Smirnov  Test (KST) to evaluate the density estimate of numerical columns, and the Total Variation Distance (TVD) for categorical columns. When calculating the correlation between columns, we employ Pearson correlation for numerical columns and contingency similarity for categorical columns. 
The results for both density estimate similarity and correlation based analysis are detailed in Table~\ref{tab:column_wise_score}. As shown,
\ourmethod synthetic data closely matches the real data in terms of univariate distribution and bivariate correlation, outperforming the baseline.

\begin{table}[htb!]
    \centering
    \caption{Low-order statistics~\cite{zhang2023mixed} of column-wise data density and pair-wise column correlation\footref{foot:sdm}. Scores range from 0 to 1. Higher values indicate more accurate estimation. \ourmethod outperforms the best generative baseline model  in most case.
    The best results are marked in \textbf{bold}, the second-best results are \underline{underlined}.
    Results are averaged over five random runs.}
    \label{tab:column_wise_score}
    \setlength{\tabcolsep}{0.5mm}
    \begin{tabular}{l|ccc|cc}
        \toprule
        \multicolumn{6}{c}{\textbf{Single-Column Shape Score ($\uparrow$)~}} \\
        \midrule
         Dataset & CTGAN & Co.GAN & TabSyn & GReaT & \textbf{\ourmethod} \\
         \midrule
         Adult & 0.81 &  \underline{0.92} & \bf 0.98 & 0.88 & 0.90 \\
         \midrule
         Beijing & 0.89 & 0.79 & \bf 0.98 & 0.93 & \underline{0.97} \\
         \midrule
         California & 0.87 & 0.77 & \bf 0.98 & \underline{0.89} & 0.83 \\
         \midrule
         US-locations & 0.83 & 0.82 & \underline{0.96} & 0.93 & \bf 0.97 \\
         \midrule
         Seattle & 0.83 & 0.73 & 0.93 & 0.90 & \bf 0.93 \\
         \midrule
         Travel & 0.84 & 0.90 & 0.93 & \bf 0.93 & \bf 0.93 \\
         \bottomrule
         \toprule
        \multicolumn{6}{c}{\textbf{Two-Column Pair Trends score ($\uparrow$)~}} \\
        \midrule
         Dataset & CTGAN & Co.GAN & TabSyn & GReaT & \textbf{\ourmethod} \\
         \midrule
         Adult & 0.81 & \underline{0.86} & \bf 0.93 & 0.80 & 0.78 \\
         \midrule
         Beijing & 0.92 & 0.94 & \bf 0.99 & 0.95 & \underline{0.98} \\
         \midrule
         California & 0.84 & 0.87 & \bf 0.97 & 0.87 & \underline{0.91} \\
         \midrule
         US-locations & 0.50 & 0.55 & \underline{0.93} & 0.89 & \bf 0.94 \\
         \midrule
         Seattle & 0.74 & 0.72 & \underline{0.80} & 0.76 & \bf 0.81 \\
         \midrule
         Travel & 0.77 & 0.80 & \bf 0.87 & \underline{0.85} & 0.82 \\
         \bottomrule
    \end{tabular}
\end{table}

\begin{figure}[htp]
    \centering
    \includegraphics[width=\linewidth]{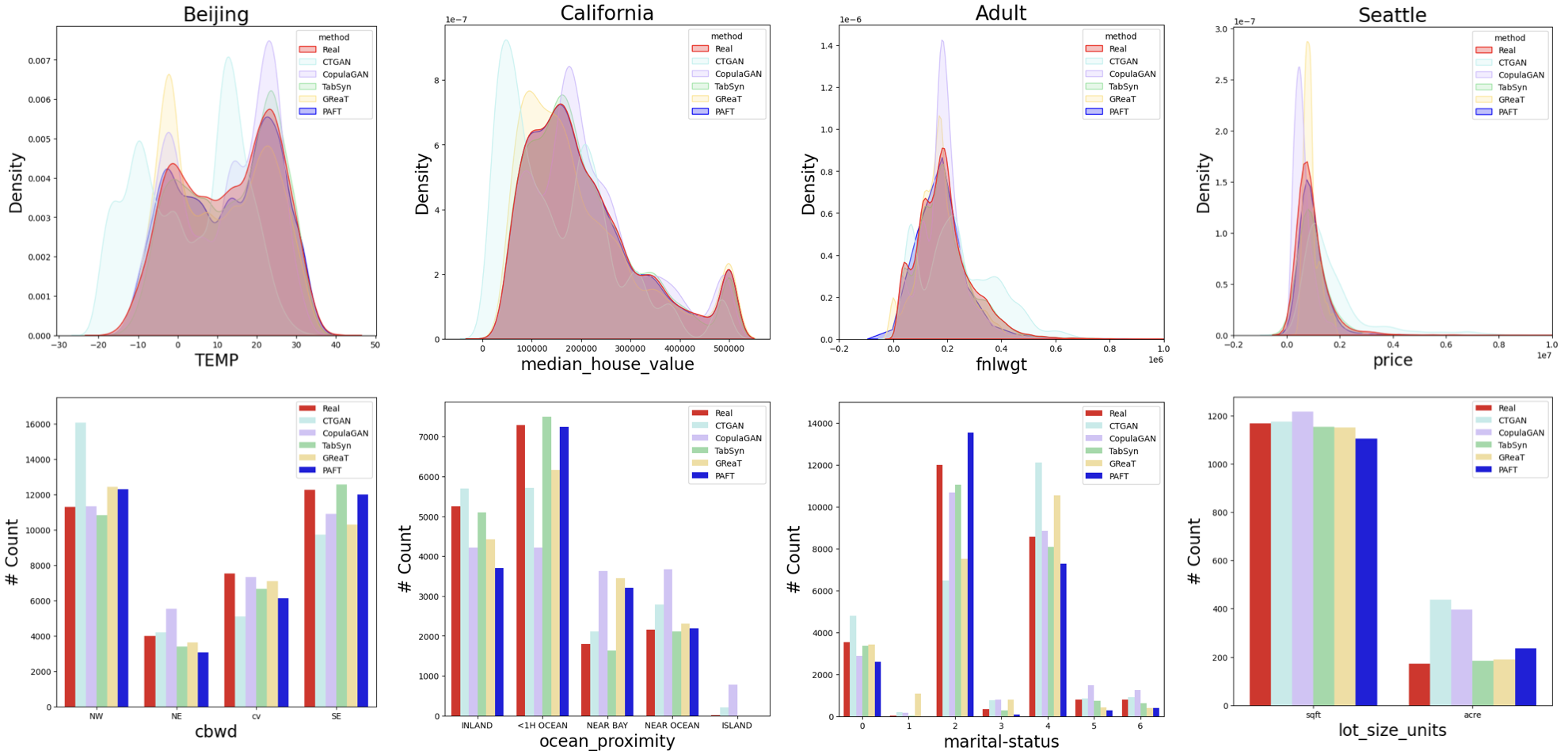}
    \caption{Column distributions visualization for each dataset generated by CTGAN, CopulaGAN, GReaT, and \ourmethod. The top row displays examples of numerical columns, while the bottom row presents examples of categorical columns.
    Overall, \ourmethod (Blue) has the closest distribution to real data (Red) compared to other synthesis methods. \ourmethod also showcase the ability to generate a wide range of diversity.}
    \label{fig:distribution_plots_partial}
\end{figure}

Visual examples are depicted in Figure~\ref{fig:distribution_plots_partial}. \ourmethod has the ability to generate a wide range of diversity, encompassing both continuous and discrete variables, which closely resembles real data.

\subsection{RQ6: Do newer generations of LLMs obviate the need for \ourmethod{}?}
\label{sec:discussion_gpt4}

The choice of foundation models for synthetic table generation involves two key considerations: in-context learning and fine-tuning, especially with large-parameter models such as GPT-4 and LLaMA. A recent survey~\cite{fang2024large} indicates that the in-context learning approach of GPT-4 is well-suited for augmenting tabular data in low-data regimes, as highlighted by CLLM~\cite{cllm2024}. However, GPT-4 suffers from limitations such as the disappearance of column-wise tail distributions and a low success rate in accurately extracting output cell values~\cite{fang2024large}.

In contrast, the state-of-the-art GReaT~\cite{borisov2022language} and other synthetic data generation approaches~\cite{solatorio2023realtabformer,zhang2023generative,zhao2023tabula} typically employ smaller models like GPT-2 or DistilGPT2, which effectively address attribute encoding while reducing feature engineering efforts. These smaller models already outperform traditional GAN and traditional statistics-based approaches. For example, DistilGPT2 can be trained using LoRA on GPUs as modest as the Tesla P100.

Importantly, fine-tuning larger models such as GPT-4 entails significant computational cost. For instance, fine-tuning a GPT-4 model for three epochs on a dataset with 50k rows and a five-column table costs approximately \$300 (OpenAI) or requires equivalent high-end GPU resources. Therefore, while the same fine-tuning schema can be applied to models like GPT-4 or LLaMA, it is not a cost-effective solution compared to the solutions considered here.

\textbf{In summary, newer generation LLMs do not obviate the need for \ourmethod{}. Our method directly addresses the impedance mismatch between autoregressive LLMs and synthetic table generation—particularly by preserving functional dependencies through permutation-aware fine-tuning—a challenge that persists even with advanced models.}

\section{Conclusion}
\label{sec:conclusion}

This work has brought LLMs closer to the goal of generating realistic synthetic datasets. By learning FDs and leveraging this information in the fine-tuning process, we are able to align the auto-regressive nature of LLMs with the ordering of columns necessary for generating quality synthetic data. While \ourmethod
is quite broadly applicable by itself, it can be extended in several directions. First, what are other, perhaps more expressive, types of tabular constraints that can be utilized in the fine-tuning process? Second, what is the internal basis for regulating orders inside a transformer architecture and can we more directly harness it? Third, can we theoretically prove the (im)possibility of generating specific synthetic datasets by LLM architectures? Fourth, despite the numerous advantages of LLM in learning and generating tabular data, scalability remains an acknowledged challenge~\cite{dinh2022lift,borisov2022language,fang2024large}, encompassing concerns such as context window and training speed. And finally, privacy-preserving methods have been implemented in table generators based on GANs but
remains understudied in LLM fine-tuning. These questions will be the focus of our future work.

\textbf{Limitations}. 
The row-wise generation cost of our method, particularly when employing fine-tuning, is affected by the dataset sample size and computational resources (GPU). Moreover, the capacity of \ourmethod to generate columns is affected by context window sizes.
These limitations can be overcome by the newer generation of LLMs or by exploring partial row generation, i.e.,
generating a row in multiple steps using an LLM.



\bibliographystyle{ACM-Reference-Format}
\bibliography{reference}


\begin{thebibliography}{61}


\ifx \showCODEN    \undefined \def \showCODEN     #1{\unskip}     \fi
\ifx \showISBNx    \undefined \def \showISBNx     #1{\unskip}     \fi
\ifx \showISBNxiii \undefined \def \showISBNxiii  #1{\unskip}     \fi
\ifx \showISSN     \undefined \def \showISSN      #1{\unskip}     \fi
\ifx \showLCCN     \undefined \def \showLCCN      #1{\unskip}     \fi
\ifx \shownote     \undefined \def \shownote      #1{#1}          \fi
\ifx \showarticletitle \undefined \def \showarticletitle #1{#1}   \fi
\ifx \showURL      \undefined \def \showURL       {\relax}        \fi
\providecommand\bibfield[2]{#2}
\providecommand\bibinfo[2]{#2}
\providecommand\natexlab[1]{#1}
\providecommand\showeprint[2][]{arXiv:#2}

\bibitem[Achiam et~al\mbox{.}(2023)]%
        {achiam2023gpt}
\bibfield{author}{\bibinfo{person}{Josh Achiam}, \bibinfo{person}{Steven Adler}, \bibinfo{person}{Sandhini Agarwal}, \bibinfo{person}{Lama Ahmad}, \bibinfo{person}{Ilge Akkaya}, \bibinfo{person}{Florencia~Leoni Aleman}, \bibinfo{person}{Diogo Almeida}, \bibinfo{person}{Janko Altenschmidt}, \bibinfo{person}{Sam Altman}, \bibinfo{person}{Shyamal Anadkat}, {et~al\mbox{.}}} \bibinfo{year}{2023}\natexlab{}.
\newblock \showarticletitle{Gpt-4 technical report}.
\newblock \bibinfo{journal}{\emph{arXiv preprint arXiv:2303.08774}} (\bibinfo{year}{2023}).
\newblock


\bibitem[Barry~Becker(1996)]%
        {adult}
\bibfield{author}{\bibinfo{person}{Ronny~Kohavi Barry~Becker}.} \bibinfo{year}{1996}\natexlab{}.
\newblock \bibinfo{title}{Adult}.
\newblock
\urldef\tempurl%
\url{https://archive.ics.uci.edu/dataset/2/adult}
\showURL{%
\tempurl}


\bibitem[Borisov et~al\mbox{.}(2022)]%
        {borisov2022language}
\bibfield{author}{\bibinfo{person}{Vadim Borisov}, \bibinfo{person}{Kathrin Se{\ss}ler}, \bibinfo{person}{Tobias Leemann}, \bibinfo{person}{Martin Pawelczyk}, {and} \bibinfo{person}{Gjergji Kasneci}.} \bibinfo{year}{2022}\natexlab{}.
\newblock \showarticletitle{Language models are realistic tabular data generators}.
\newblock \bibinfo{journal}{\emph{arXiv preprint arXiv:2210.06280}} (\bibinfo{year}{2022}).
\newblock


\bibitem[Brown et~al\mbox{.}(2020)]%
        {brown2020language}
\bibfield{author}{\bibinfo{person}{Tom Brown}, \bibinfo{person}{Benjamin Mann}, \bibinfo{person}{Nick Ryder}, \bibinfo{person}{Melanie Subbiah}, \bibinfo{person}{Jared~D Kaplan}, \bibinfo{person}{Prafulla Dhariwal}, \bibinfo{person}{Arvind Neelakantan}, \bibinfo{person}{Pranav Shyam}, \bibinfo{person}{Girish Sastry}, \bibinfo{person}{Amanda Askell}, {et~al\mbox{.}}} \bibinfo{year}{2020}\natexlab{}.
\newblock \showarticletitle{Language models are few-shot learners}.
\newblock \bibinfo{journal}{\emph{Advances in neural information processing systems}}  \bibinfo{volume}{33} (\bibinfo{year}{2020}), \bibinfo{pages}{1877--1901}.
\newblock


\bibitem[Chen et~al\mbox{.}(2019)]%
        {chen2019faketables}
\bibfield{author}{\bibinfo{person}{Haipeng Chen}, \bibinfo{person}{Sushil Jajodia}, \bibinfo{person}{Jing Liu}, \bibinfo{person}{Noseong Park}, \bibinfo{person}{Vadim Sokolov}, {and} \bibinfo{person}{VS Subrahmanian}.} \bibinfo{year}{2019}\natexlab{}.
\newblock \showarticletitle{FakeTables: Using GANs to Generate Functional Dependency Preserving Tables with Bounded Real Data.}. In \bibinfo{booktitle}{\emph{IJCAI}}. \bibinfo{pages}{2074--2080}.
\newblock


\bibitem[Chen et~al\mbox{.}(2023)]%
        {chen2023graph}
\bibfield{author}{\bibinfo{person}{Hongjie Chen}, \bibinfo{person}{Ryan~A Rossi}, \bibinfo{person}{Kanak Mahadik}, \bibinfo{person}{Sungchul Kim}, {and} \bibinfo{person}{Hoda Eldardiry}.} \bibinfo{year}{2023}\natexlab{}.
\newblock \showarticletitle{Graph Deep Factors for Probabilistic Time-series Forecasting}.
\newblock \bibinfo{journal}{\emph{ACM Transactions on Knowledge Discovery from Data}} \bibinfo{volume}{17}, \bibinfo{number}{2} (\bibinfo{year}{2023}), \bibinfo{pages}{1--30}.
\newblock


\bibitem[Chen(2017)]%
        {beijing}
\bibfield{author}{\bibinfo{person}{Song Chen}.} \bibinfo{year}{2017}\natexlab{}.
\newblock \bibinfo{title}{Beijing PM2.5 Data}.
\newblock
\urldef\tempurl%
\url{https://archive.ics.uci.edu/dataset/381/beijing+pm2+5+data}
\showURL{%
\tempurl}


\bibitem[Chen et~al\mbox{.}(2024)]%
        {chen2024premise}
\bibfield{author}{\bibinfo{person}{Xinyun Chen}, \bibinfo{person}{Ryan~A Chi}, \bibinfo{person}{Xuezhi Wang}, {and} \bibinfo{person}{Denny Zhou}.} \bibinfo{year}{2024}\natexlab{}.
\newblock \showarticletitle{Premise Order Matters in Reasoning with Large Language Models}.
\newblock \bibinfo{journal}{\emph{arXiv preprint arXiv:2402.08939}} (\bibinfo{year}{2024}).
\newblock


\bibitem[Cvitkovic(2020)]%
        {cvitkovic2020supervised}
\bibfield{author}{\bibinfo{person}{Milan Cvitkovic}.} \bibinfo{year}{2020}\natexlab{}.
\newblock \showarticletitle{Supervised learning on relational databases with graph neural networks}.
\newblock \bibinfo{journal}{\emph{arXiv preprint arXiv:2002.02046}} (\bibinfo{year}{2020}).
\newblock


\bibitem[Dinh et~al\mbox{.}(2022)]%
        {dinh2022lift}
\bibfield{author}{\bibinfo{person}{Tuan Dinh}, \bibinfo{person}{Yuchen Zeng}, \bibinfo{person}{Ruisu Zhang}, \bibinfo{person}{Ziqian Lin}, \bibinfo{person}{Michael Gira}, \bibinfo{person}{Shashank Rajput}, \bibinfo{person}{Jy-yong Sohn}, \bibinfo{person}{Dimitris Papailiopoulos}, {and} \bibinfo{person}{Kangwook Lee}.} \bibinfo{year}{2022}\natexlab{}.
\newblock \showarticletitle{Lift: Language-interfaced fine-tuning for non-language machine learning tasks}.
\newblock \bibinfo{journal}{\emph{Advances in Neural Information Processing Systems}}  \bibinfo{volume}{35} (\bibinfo{year}{2022}), \bibinfo{pages}{11763--11784}.
\newblock


\bibitem[Distiawan et~al\mbox{.}(2018)]%
        {gtr_lstm}
\bibfield{author}{\bibinfo{person}{Bayu Distiawan}, \bibinfo{person}{Jianzhong Qi}, \bibinfo{person}{Rui Zhang}, {and} \bibinfo{person}{Wei Wang}.} \bibinfo{year}{2018}\natexlab{}.
\newblock \showarticletitle{GTR-LSTM: A triple encoder for sentence generation from RDF data}. In \bibinfo{booktitle}{\emph{Proceedings of the 56th Annual Meeting of the Association for Computational Linguistics (Volume 1: Long Papers)}}. \bibinfo{pages}{1627--1637}.
\newblock


\bibitem[Du et~al\mbox{.}(2021)]%
        {du2021tabularnet}
\bibfield{author}{\bibinfo{person}{Lun Du}, \bibinfo{person}{Fei Gao}, \bibinfo{person}{Xu Chen}, \bibinfo{person}{Ran Jia}, \bibinfo{person}{Junshan Wang}, \bibinfo{person}{Jiang Zhang}, \bibinfo{person}{Shi Han}, {and} \bibinfo{person}{Dongmei Zhang}.} \bibinfo{year}{2021}\natexlab{}.
\newblock \showarticletitle{TabularNet: A neural network architecture for understanding semantic structures of tabular data}. In \bibinfo{booktitle}{\emph{Proceedings of the 27th ACM SIGKDD Conference on Knowledge Discovery \& Data Mining}}. \bibinfo{pages}{322--331}.
\newblock


\bibitem[Du and Li(2024)]%
        {du2024towards}
\bibfield{author}{\bibinfo{person}{Yuntao Du} {and} \bibinfo{person}{Ninghui Li}.} \bibinfo{year}{2024}\natexlab{}.
\newblock \showarticletitle{Towards principled assessment of tabular data synthesis algorithms}.
\newblock \bibinfo{journal}{\emph{arXiv preprint arXiv:2402.06806}} (\bibinfo{year}{2024}).
\newblock


\bibitem[Evans-Bye(2015)]%
        {arcgisStatesShapefile}
\bibfield{author}{\bibinfo{person}{Dominique Evans-Bye}.} \bibinfo{year}{2015}\natexlab{}.
\newblock \bibinfo{title}{States Shapefile}.
\newblock \bibinfo{howpublished}{\url{https://hub.arcgis.com/datasets/CMHS::states-shapefile/explore?location=29.721532\%2C71.941464\%2C3.76}}.
\newblock


\bibitem[Fang et~al\mbox{.}(2024)]%
        {fang2024large}
\bibfield{author}{\bibinfo{person}{Xi Fang}, \bibinfo{person}{Weijie Xu}, \bibinfo{person}{Fiona~Anting Tan}, \bibinfo{person}{Jiani Zhang}, \bibinfo{person}{Ziqing Hu}, \bibinfo{person}{Yanjun~(Jane) Qi}, \bibinfo{person}{Scott Nickleach}, \bibinfo{person}{Diego Socolinsky}, \bibinfo{person}{"SHS" Srinivasan~Sengamedu}, {and} \bibinfo{person}{Christos Faloutsos}.} \bibinfo{year}{2024}\natexlab{}.
\newblock \showarticletitle{Large language models (LLMs) on tabular data: Prediction, generation, and understanding — a survey}.
\newblock \bibinfo{journal}{\emph{Transactions on Machine Learning Research}} (\bibinfo{year}{2024}).
\newblock


\bibitem[Garcia-Molina et~al\mbox{.}(2008)]%
        {10.5555/1450931}
\bibfield{author}{\bibinfo{person}{Hector Garcia-Molina}, \bibinfo{person}{Jeffrey~D. Ullman}, {and} \bibinfo{person}{Jennifer Widom}.} \bibinfo{year}{2008}\natexlab{}.
\newblock \bibinfo{booktitle}{\emph{Database Systems: The Complete Book} (\bibinfo{edition}{2} ed.)}.
\newblock \bibinfo{publisher}{Prentice Hall Press}, \bibinfo{address}{USA}.
\newblock
\showISBNx{9780131873254}


\bibitem[Guo et~al\mbox{.}(2025)]%
        {guo2025deepseek}
\bibfield{author}{\bibinfo{person}{Daya Guo}, \bibinfo{person}{Dejian Yang}, \bibinfo{person}{Haowei Zhang}, \bibinfo{person}{Junxiao Song}, \bibinfo{person}{Ruoyu Zhang}, \bibinfo{person}{Runxin Xu}, \bibinfo{person}{Qihao Zhu}, \bibinfo{person}{Shirong Ma}, \bibinfo{person}{Peiyi Wang}, \bibinfo{person}{Xiao Bi}, {et~al\mbox{.}}} \bibinfo{year}{2025}\natexlab{}.
\newblock \showarticletitle{Deepseek-r1: Incentivizing reasoning capability in llms via reinforcement learning}.
\newblock \bibinfo{journal}{\emph{arXiv preprint arXiv:2501.12948}} (\bibinfo{year}{2025}).
\newblock


\bibitem[Hoffmann et~al\mbox{.}(2022)]%
        {chinchilla_scaling_law}
\bibfield{author}{\bibinfo{person}{Jordan Hoffmann}, \bibinfo{person}{Sebastian Borgeaud}, \bibinfo{person}{Arthur Mensch}, \bibinfo{person}{Elena Buchatskaya}, \bibinfo{person}{Trevor Cai}, \bibinfo{person}{Eliza Rutherford}, \bibinfo{person}{Diego de Las~Casas}, \bibinfo{person}{Lisa~Anne Hendricks}, \bibinfo{person}{Johannes Welbl}, \bibinfo{person}{Aidan Clark}, \bibinfo{person}{Tom Hennigan}, \bibinfo{person}{Eric Noland}, \bibinfo{person}{Katie Millican}, \bibinfo{person}{George van~den Driessche}, \bibinfo{person}{Bogdan Damoc}, \bibinfo{person}{Aurelia Guy}, \bibinfo{person}{Simon Osindero}, \bibinfo{person}{Karen Simonyan}, \bibinfo{person}{Erich Elsen}, \bibinfo{person}{Jack~W. Rae}, \bibinfo{person}{Oriol Vinyals}, {and} \bibinfo{person}{Laurent Sifre}.} \bibinfo{year}{2022}\natexlab{}.
\newblock \showarticletitle{Training Compute-Optimal Large Language Models}.
\newblock \bibinfo{journal}{\emph{CoRR}}  \bibinfo{volume}{abs/2203.15556} (\bibinfo{year}{2022}).
\newblock
\href{https://doi.org/10.48550/ARXIV.2203.15556}{doi:\nolinkurl{10.48550/ARXIV.2203.15556}}
\showeprint[arXiv]{2203.15556}


\bibitem[Hu et~al\mbox{.}(2021)]%
        {LoRA}
\bibfield{author}{\bibinfo{person}{Edward~J. Hu}, \bibinfo{person}{Yelong Shen}, \bibinfo{person}{Phillip Wallis}, \bibinfo{person}{Zeyuan Allen{-}Zhu}, \bibinfo{person}{Yuanzhi Li}, \bibinfo{person}{Shean Wang}, {and} \bibinfo{person}{Weizhu Chen}.} \bibinfo{year}{2021}\natexlab{}.
\newblock \showarticletitle{LoRA: Low-Rank Adaptation of Large Language Models}.
\newblock \bibinfo{journal}{\emph{CoRR}}  \bibinfo{volume}{abs/2106.09685} (\bibinfo{year}{2021}).
\newblock
\showeprint[arXiv]{2106.09685}
\urldef\tempurl%
\url{https://arxiv.org/abs/2106.09685}
\showURL{%
\tempurl}


\bibitem[Kanter and Veeramachaneni(2015)]%
        {kanter2015deep}
\bibfield{author}{\bibinfo{person}{James~Max Kanter} {and} \bibinfo{person}{Kalyan Veeramachaneni}.} \bibinfo{year}{2015}\natexlab{}.
\newblock \showarticletitle{Deep feature synthesis: Towards automating data science endeavors}. In \bibinfo{booktitle}{\emph{2015 IEEE international conference on data science and advanced analytics (DSAA)}}. IEEE, \bibinfo{pages}{1--10}.
\newblock


\bibitem[Koller and Friedman(2009)]%
        {koller2009probabilistic}
\bibfield{author}{\bibinfo{person}{Daphne Koller} {and} \bibinfo{person}{Nir Friedman}.} \bibinfo{year}{2009}\natexlab{}.
\newblock \bibinfo{booktitle}{\emph{Probabilistic graphical models: principles and techniques}}.
\newblock \bibinfo{publisher}{MIT press}.
\newblock


\bibitem[Liang et~al\mbox{.}(2022)]%
        {liang2022holistic}
\bibfield{author}{\bibinfo{person}{Percy Liang}, \bibinfo{person}{Rishi Bommasani}, \bibinfo{person}{Tony Lee}, \bibinfo{person}{Dimitris Tsipras}, \bibinfo{person}{Dilara Soylu}, \bibinfo{person}{Michihiro Yasunaga}, \bibinfo{person}{Yian Zhang}, \bibinfo{person}{Deepak Narayanan}, \bibinfo{person}{Yuhuai Wu}, \bibinfo{person}{Ananya Kumar}, {et~al\mbox{.}}} \bibinfo{year}{2022}\natexlab{}.
\newblock \showarticletitle{Holistic evaluation of language models}.
\newblock \bibinfo{journal}{\emph{arXiv preprint arXiv:2211.09110}} (\bibinfo{year}{2022}).
\newblock


\bibitem[Lin et~al\mbox{.}(2020)]%
        {lin2020using}
\bibfield{author}{\bibinfo{person}{Zinan Lin}, \bibinfo{person}{Alankar Jain}, \bibinfo{person}{Chen Wang}, \bibinfo{person}{Giulia Fanti}, {and} \bibinfo{person}{Vyas Sekar}.} \bibinfo{year}{2020}\natexlab{}.
\newblock \showarticletitle{Using gans for sharing networked time series data: Challenges, initial promise, and open questions}. In \bibinfo{booktitle}{\emph{Proceedings of the ACM Internet Measurement Conference}}. \bibinfo{pages}{464--483}.
\newblock


\bibitem[Liu et~al\mbox{.}(2024)]%
        {liu2024deepseek}
\bibfield{author}{\bibinfo{person}{Aixin Liu}, \bibinfo{person}{Bei Feng}, \bibinfo{person}{Bing Xue}, \bibinfo{person}{Bingxuan Wang}, \bibinfo{person}{Bochao Wu}, \bibinfo{person}{Chengda Lu}, \bibinfo{person}{Chenggang Zhao}, \bibinfo{person}{Chengqi Deng}, \bibinfo{person}{Chenyu Zhang}, \bibinfo{person}{Chong Ruan}, {et~al\mbox{.}}} \bibinfo{year}{2024}\natexlab{}.
\newblock \showarticletitle{Deepseek-v3 technical report}.
\newblock \bibinfo{journal}{\emph{arXiv preprint arXiv:2412.19437}} (\bibinfo{year}{2024}).
\newblock


\bibitem[Liu et~al\mbox{.}(2022)]%
        {liu2022goggle}
\bibfield{author}{\bibinfo{person}{Tennison Liu}, \bibinfo{person}{Zhaozhi Qian}, \bibinfo{person}{Jeroen Berrevoets}, {and} \bibinfo{person}{Mihaela van~der Schaar}.} \bibinfo{year}{2022}\natexlab{}.
\newblock \showarticletitle{GOGGLE: Generative modelling for tabular data by learning relational structure}. In \bibinfo{booktitle}{\emph{The Eleventh International Conference on Learning Representations}}.
\newblock


\bibitem[Mandros et~al\mbox{.}(2020)]%
        {mandros2020discovering}
\bibfield{author}{\bibinfo{person}{Panagiotis Mandros}, \bibinfo{person}{David Kaltenpoth}, \bibinfo{person}{Mario Boley}, {and} \bibinfo{person}{Jilles Vreeken}.} \bibinfo{year}{2020}\natexlab{}.
\newblock \showarticletitle{Discovering functional dependencies from mixed-type data}. In \bibinfo{booktitle}{\emph{Proceedings of the 26th ACM SIGKDD International Conference on Knowledge Discovery \& Data Mining}}. \bibinfo{pages}{1404--1414}.
\newblock


\bibitem[Margeloiu et~al\mbox{.}({[n.\,d.]})]%
        {margeloiutabebm}
\bibfield{author}{\bibinfo{person}{Andrei Margeloiu}, \bibinfo{person}{Xiangjian Jiang}, \bibinfo{person}{Nikola Simidjievski}, {and} \bibinfo{person}{Mateja Jamnik}.} \bibinfo{year}{[n.\,d.]}\natexlab{}.
\newblock \showarticletitle{TabEBM: A Tabular Data Augmentation Method with Distinct Class-Specific Energy-Based Models}. In \bibinfo{booktitle}{\emph{The Thirty-eighth Annual Conference on Neural Information Processing Systems}}.
\newblock


\bibitem[McKenna et~al\mbox{.}(2019)]%
        {mckenna2019graphical}
\bibfield{author}{\bibinfo{person}{Ryan McKenna}, \bibinfo{person}{Daniel Sheldon}, {and} \bibinfo{person}{Gerome Miklau}.} \bibinfo{year}{2019}\natexlab{}.
\newblock \showarticletitle{Graphical-model based estimation and inference for differential privacy}. In \bibinfo{booktitle}{\emph{International Conference on Machine Learning}}. PMLR, \bibinfo{pages}{4435--4444}.
\newblock


\bibitem[Miller and Blair(2009)]%
        {miller2009input}
\bibfield{author}{\bibinfo{person}{R.~E. Miller} {and} \bibinfo{person}{P.~D. Blair}.} \bibinfo{year}{2009}\natexlab{}.
\newblock \bibinfo{booktitle}{\emph{Input-output analysis: foundations and extensions}}.
\newblock \bibinfo{publisher}{Cambridge university press}.
\newblock


\bibitem[Muralidhar et~al\mbox{.}(2018)]%
        {muralidhar2018illiad}
\bibfield{author}{\bibinfo{person}{Nikhil Muralidhar}, \bibinfo{person}{Chen Wang}, \bibinfo{person}{Nathan Self}, \bibinfo{person}{Marjan Momtazpour}, \bibinfo{person}{Kiyoshi Nakayama}, \bibinfo{person}{Ratnesh Sharma}, {and} \bibinfo{person}{Naren Ramakrishnan}.} \bibinfo{year}{2018}\natexlab{}.
\newblock \showarticletitle{illiad: Intelligent invariant and anomaly detection in cyber-physical systems}.
\newblock \bibinfo{journal}{\emph{ACM Transactions on Intelligent Systems and Technology (TIST)}} \bibinfo{volume}{9}, \bibinfo{number}{3} (\bibinfo{year}{2018}), \bibinfo{pages}{1--20}.
\newblock


\bibitem[Nugent(2017)]%
        {california}
\bibfield{author}{\bibinfo{person}{Cam Nugent}.} \bibinfo{year}{2017}\natexlab{}.
\newblock \bibinfo{title}{California Housing Prices}.
\newblock
\urldef\tempurl%
\url{https://www.kaggle.com/datasets/camnugent/california-housing-prices}
\showURL{%
\tempurl}


\bibitem[OpenAI(2022)]%
        {chatgpt}
\bibfield{author}{\bibinfo{person}{OpenAI}.} \bibinfo{year}{2022}\natexlab{}.
\newblock \bibinfo{title}{Chat-GPT: Optimizing Language Models for Dialogue}.
\newblock
\urldef\tempurl%
\url{https://openai.com/blog/chatgpt/}
\showURL{%
\tempurl}


\bibitem[Papenbrock et~al\mbox{.}(2015)]%
        {papenbrock2015functional}
\bibfield{author}{\bibinfo{person}{Thorsten Papenbrock}, \bibinfo{person}{Jens Ehrlich}, \bibinfo{person}{Jannik Marten}, \bibinfo{person}{Tommy Neubert}, \bibinfo{person}{Jan-Peer Rudolph}, \bibinfo{person}{Martin Sch{\"o}nberg}, \bibinfo{person}{Jakob Zwiener}, {and} \bibinfo{person}{Felix Naumann}.} \bibinfo{year}{2015}\natexlab{}.
\newblock \showarticletitle{Functional dependency discovery: An experimental evaluation of seven algorithms}.
\newblock \bibinfo{journal}{\emph{Proceedings of the VLDB Endowment}} \bibinfo{volume}{8}, \bibinfo{number}{10} (\bibinfo{year}{2015}), \bibinfo{pages}{1082--1093}.
\newblock


\bibitem[Papenbrock and Naumann(2016)]%
        {papenbrock2016hybrid}
\bibfield{author}{\bibinfo{person}{Thorsten Papenbrock} {and} \bibinfo{person}{Felix Naumann}.} \bibinfo{year}{2016}\natexlab{}.
\newblock \showarticletitle{A hybrid approach to functional dependency discovery}. In \bibinfo{booktitle}{\emph{Proceedings of the 2016 International Conference on Management of Data}}. \bibinfo{pages}{821--833}.
\newblock


\bibitem[Park et~al\mbox{.}(2018)]%
        {park2018data}
\bibfield{author}{\bibinfo{person}{Noseong Park}, \bibinfo{person}{Mahmoud Mohammadi}, \bibinfo{person}{Kshitij Gorde}, \bibinfo{person}{Sushil Jajodia}, \bibinfo{person}{Hongkyu Park}, {and} \bibinfo{person}{Youngmin Kim}.} \bibinfo{year}{2018}\natexlab{}.
\newblock \showarticletitle{Data synthesis based on generative adversarial networks}.
\newblock \bibinfo{journal}{\emph{Proceedings of the VLDB Endowment}} \bibinfo{volume}{11}, \bibinfo{number}{10} (\bibinfo{year}{2018}), \bibinfo{pages}{1071--1083}.
\newblock


\bibitem[Patki et~al\mbox{.}(2016)]%
        {SDV}
\bibfield{author}{\bibinfo{person}{Neha Patki}, \bibinfo{person}{Roy Wedge}, {and} \bibinfo{person}{Kalyan Veeramachaneni}.} \bibinfo{year}{2016}\natexlab{}.
\newblock \showarticletitle{The Synthetic data vault}. In \bibinfo{booktitle}{\emph{IEEE International Conference on Data Science and Advanced Analytics (DSAA)}}. \bibinfo{pages}{399--410}.
\newblock
\href{https://doi.org/10.1109/DSAA.2016.49}{doi:\nolinkurl{10.1109/DSAA.2016.49}}


\bibitem[Pennerath et~al\mbox{.}(2020)]%
        {pennerath2020discovering}
\bibfield{author}{\bibinfo{person}{Fr{\'e}d{\'e}ric Pennerath}, \bibinfo{person}{Panagiotis Mandros}, {and} \bibinfo{person}{Jilles Vreeken}.} \bibinfo{year}{2020}\natexlab{}.
\newblock \showarticletitle{Discovering approximate functional dependencies using smoothed mutual information}. In \bibinfo{booktitle}{\emph{Proceedings of the 26th ACM SIGKDD International Conference on Knowledge Discovery \& Data Mining}}. \bibinfo{pages}{1254--1264}.
\newblock


\bibitem[Prystawski et~al\mbox{.}(2024)]%
        {prystawski2024think}
\bibfield{author}{\bibinfo{person}{Ben Prystawski}, \bibinfo{person}{Michael Li}, {and} \bibinfo{person}{Noah Goodman}.} \bibinfo{year}{2024}\natexlab{}.
\newblock \showarticletitle{Why think step by step? Reasoning emerges from the locality of experience}.
\newblock \bibinfo{journal}{\emph{Advances in Neural Information Processing Systems}}  \bibinfo{volume}{36} (\bibinfo{year}{2024}).
\newblock


\bibitem[Radford et~al\mbox{.}(2019)]%
        {radford2019language}
\bibfield{author}{\bibinfo{person}{Alec Radford}, \bibinfo{person}{Jeffrey Wu}, \bibinfo{person}{Rewon Child}, \bibinfo{person}{David Luan}, \bibinfo{person}{Dario Amodei}, \bibinfo{person}{Ilya Sutskever}, {et~al\mbox{.}}} \bibinfo{year}{2019}\natexlab{}.
\newblock \showarticletitle{Language models are unsupervised multitask learners}.
\newblock \bibinfo{journal}{\emph{OpenAI blog}} \bibinfo{volume}{1}, \bibinfo{number}{8} (\bibinfo{year}{2019}), \bibinfo{pages}{9}.
\newblock


\bibitem[Raffel et~al\mbox{.}(2020)]%
        {raffel:20}
\bibfield{author}{\bibinfo{person}{C. Raffel}, \bibinfo{person}{N. Shazeer}, \bibinfo{person}{A. Roberts}, \bibinfo{person}{K. Lee}, \bibinfo{person}{Sharan S.~Narang}, \bibinfo{person}{M. Matena}, \bibinfo{person}{Y. Zhou}, \bibinfo{person}{W. Li}, {and} \bibinfo{person}{P.~J. Liu}.} \bibinfo{year}{2020}\natexlab{}.
\newblock \showarticletitle{Exploring the Limits of Transfer Learning with a Unified Text-to-Text Transformer}.
\newblock \bibinfo{journal}{\emph{Journal of Machine Learning Research}}  \bibinfo{volume}{21} (\bibinfo{year}{2020}), \bibinfo{pages}{1--67}.
\newblock


\bibitem[Ribeiro et~al\mbox{.}(2019)]%
        {dual_graph}
\bibfield{author}{\bibinfo{person}{Leonardo~FR Ribeiro}, \bibinfo{person}{Claire Gardent}, {and} \bibinfo{person}{Iryna Gurevych}.} \bibinfo{year}{2019}\natexlab{}.
\newblock \showarticletitle{Enhancing AMR-to-Text Generation with Dual Graph Representations}. In \bibinfo{booktitle}{\emph{Proceedings of the 2019 Conference on Empirical Methods in Natural Language Processing and the 9th International Joint Conference on Natural Language Processing (EMNLP-IJCNLP)}}. \bibinfo{pages}{3183--3194}.
\newblock


\bibitem[Romera-Paredes et~al\mbox{.}(2023)]%
        {romera2023mathematical}
\bibfield{author}{\bibinfo{person}{Bernardino Romera-Paredes}, \bibinfo{person}{Mohammadamin Barekatain}, \bibinfo{person}{Alexander Novikov}, \bibinfo{person}{Matej Balog}, \bibinfo{person}{M~Pawan Kumar}, \bibinfo{person}{Emilien Dupont}, \bibinfo{person}{Francisco~JR Ruiz}, \bibinfo{person}{Jordan~S Ellenberg}, \bibinfo{person}{Pengming Wang}, \bibinfo{person}{Omar Fawzi}, {et~al\mbox{.}}} \bibinfo{year}{2023}\natexlab{}.
\newblock \showarticletitle{Mathematical discoveries from program search with large language models}.
\newblock \bibinfo{journal}{\emph{Nature}} (\bibinfo{year}{2023}), \bibinfo{pages}{1--3}.
\newblock


\bibitem[samuel Cortinhas(2022)]%
        {seattle}
\bibfield{author}{\bibinfo{person}{samuel Cortinhas}.} \bibinfo{year}{2022}\natexlab{}.
\newblock \bibinfo{title}{HOUSE PRICE PREDICTION - SEATTLE}.
\newblock
\urldef\tempurl%
\url{https://www.kaggle.com/datasets/samuelcortinhas/house-price-prediction-seattle}
\showURL{%
\tempurl}


\bibitem[Seedat et~al\mbox{.}(2024)]%
        {cllm2024}
\bibfield{author}{\bibinfo{person}{Nabeel Seedat}, \bibinfo{person}{Nicolas Huynh}, \bibinfo{person}{Boris van Breugel}, {and} \bibinfo{person}{Mihaela van~der Schaar}.} \bibinfo{year}{2024}\natexlab{}.
\newblock \showarticletitle{Curated {LLM}: Synergy of {LLM}s and Data Curation for tabular augmentation in low-data regimes}. In \bibinfo{booktitle}{\emph{Forty-first International Conference on Machine Learning}}.
\newblock


\bibitem[Sharma et~al\mbox{.}(2024)]%
        {dtg_survey}
\bibfield{author}{\bibinfo{person}{Mandar Sharma}, \bibinfo{person}{Ajay~Kumar Gogineni}, {and} \bibinfo{person}{Naren Ramakrishnan}.} \bibinfo{year}{2024}\natexlab{}.
\newblock \showarticletitle{Neural Methods for Data-to-text Generation}.
\newblock \bibinfo{journal}{\emph{ACM Transactions on Intelligent Systems and Technology}} (\bibinfo{year}{2024}).
\newblock


\bibitem[Solatorio and Dupriez(2023)]%
        {solatorio2023realtabformer}
\bibfield{author}{\bibinfo{person}{Aivin~V Solatorio} {and} \bibinfo{person}{Olivier Dupriez}.} \bibinfo{year}{2023}\natexlab{}.
\newblock \showarticletitle{Realtabformer: Generating realistic relational and tabular data using transformers}.
\newblock \bibinfo{journal}{\emph{arXiv preprint arXiv:2302.02041}} (\bibinfo{year}{2023}).
\newblock


\bibitem[Tejashvi(2021)]%
        {travel}
\bibfield{author}{\bibinfo{person}{Tejashvi}.} \bibinfo{year}{2021}\natexlab{}.
\newblock \bibinfo{title}{Tour \& Travels Customer Churn Prediction}.
\newblock
\urldef\tempurl%
\url{https://www.kaggle.com/datasets/tejashvi14/tour-travels-customer-churn-prediction}
\showURL{%
\tempurl}


\bibitem[Touvron et~al\mbox{.}(2023)]%
        {touvron2023llama}
\bibfield{author}{\bibinfo{person}{Hugo Touvron}, \bibinfo{person}{Louis Martin}, \bibinfo{person}{Kevin Stone}, \bibinfo{person}{Peter Albert}, \bibinfo{person}{Amjad Almahairi}, \bibinfo{person}{Yasmine Babaei}, \bibinfo{person}{Nikolay Bashlykov}, \bibinfo{person}{Soumya Batra}, \bibinfo{person}{Prajjwal Bhargava}, \bibinfo{person}{Shruti Bhosale}, {et~al\mbox{.}}} \bibinfo{year}{2023}\natexlab{}.
\newblock \showarticletitle{Llama 2: Open foundation and fine-tuned chat models}.
\newblock \bibinfo{journal}{\emph{arXiv preprint arXiv:2307.09288}} (\bibinfo{year}{2023}).
\newblock


\bibitem[Vaswani et~al\mbox{.}(2017)]%
        {attention_all_you_need}
\bibfield{author}{\bibinfo{person}{Ashish Vaswani}, \bibinfo{person}{Noam Shazeer}, \bibinfo{person}{Niki Parmar}, \bibinfo{person}{Jakob Uszkoreit}, \bibinfo{person}{Llion Jones}, \bibinfo{person}{Aidan~N. Gomez}, \bibinfo{person}{Lukasz Kaiser}, {and} \bibinfo{person}{Illia Polosukhin}.} \bibinfo{year}{2017}\natexlab{}.
\newblock \showarticletitle{Attention Is All You Need}.
\newblock \bibinfo{journal}{\emph{CoRR}}  \bibinfo{volume}{abs/1706.03762} (\bibinfo{year}{2017}).
\newblock
\showeprint[arXiv]{1706.03762}
\urldef\tempurl%
\url{http://arxiv.org/abs/1706.03762}
\showURL{%
\tempurl}


\bibitem[Wang et~al\mbox{.}(2020)]%
        {8805442}
\bibfield{author}{\bibinfo{person}{Yun Wang}, \bibinfo{person}{Zhida Sun}, \bibinfo{person}{Haidong Zhang}, \bibinfo{person}{Weiwei Cui}, \bibinfo{person}{Ke Xu}, \bibinfo{person}{Xiaojuan Ma}, {and} \bibinfo{person}{Dongmei Zhang}.} \bibinfo{year}{2020}\natexlab{}.
\newblock \showarticletitle{DataShot: Automatic Generation of Fact Sheets from Tabular Data}.
\newblock \bibinfo{journal}{\emph{IEEE Transactions on Visualization and Computer Graphics}} \bibinfo{volume}{26}, \bibinfo{number}{1} (\bibinfo{year}{2020}), \bibinfo{pages}{895--905}.
\newblock
\href{https://doi.org/10.1109/TVCG.2019.2934398}{doi:\nolinkurl{10.1109/TVCG.2019.2934398}}


\bibitem[Wang et~al\mbox{.}(2021)]%
        {wang2021tuta}
\bibfield{author}{\bibinfo{person}{Zhiruo Wang}, \bibinfo{person}{Haoyu Dong}, \bibinfo{person}{Ran Jia}, \bibinfo{person}{Jia Li}, \bibinfo{person}{Zhiyi Fu}, \bibinfo{person}{Shi Han}, {and} \bibinfo{person}{Dongmei Zhang}.} \bibinfo{year}{2021}\natexlab{}.
\newblock \showarticletitle{Tuta: Tree-based transformers for generally structured table pre-training}. In \bibinfo{booktitle}{\emph{Proceedings of the 27th ACM SIGKDD Conference on Knowledge Discovery \& Data Mining}}. \bibinfo{pages}{1780--1790}.
\newblock


\bibitem[Xu et~al\mbox{.}(2019)]%
        {xu2019modeling}
\bibfield{author}{\bibinfo{person}{Lei Xu}, \bibinfo{person}{Maria Skoularidou}, \bibinfo{person}{Alfredo Cuesta-Infante}, {and} \bibinfo{person}{Kalyan Veeramachaneni}.} \bibinfo{year}{2019}\natexlab{}.
\newblock \showarticletitle{Modeling tabular data using conditional gan}.
\newblock \bibinfo{journal}{\emph{Advances in neural information processing systems}}  \bibinfo{volume}{32} (\bibinfo{year}{2019}).
\newblock


\bibitem[Xu and Veeramachaneni(2018)]%
        {xu2018synthesizing}
\bibfield{author}{\bibinfo{person}{Lei Xu} {and} \bibinfo{person}{Kalyan Veeramachaneni}.} \bibinfo{year}{2018}\natexlab{}.
\newblock \showarticletitle{Synthesizing tabular data using generative adversarial networks}.
\newblock \bibinfo{journal}{\emph{arXiv preprint arXiv:1811.11264}} (\bibinfo{year}{2018}).
\newblock


\bibitem[Xu et~al\mbox{.}(2021)]%
        {xu2021stan}
\bibfield{author}{\bibinfo{person}{Shengzhe Xu}, \bibinfo{person}{Manish Marwah}, \bibinfo{person}{Martin Arlitt}, {and} \bibinfo{person}{Naren Ramakrishnan}.} \bibinfo{year}{2021}\natexlab{}.
\newblock \showarticletitle{Stan: Synthetic network traffic generation with generative neural models}. In \bibinfo{booktitle}{\emph{Deployable Machine Learning for Security Defense: Second International Workshop, MLHat 2021, Virtual Event, August 15, 2021, Proceedings 2}}. Springer, \bibinfo{pages}{3--29}.
\newblock


\bibitem[Zhang et~al\mbox{.}(2023b)]%
        {zhang2023mixed}
\bibfield{author}{\bibinfo{person}{Hengrui Zhang}, \bibinfo{person}{Jiani Zhang}, \bibinfo{person}{Balasubramaniam Srinivasan}, \bibinfo{person}{Zhengyuan Shen}, \bibinfo{person}{Xiao Qin}, \bibinfo{person}{Christos Faloutsos}, \bibinfo{person}{Huzefa Rangwala}, {and} \bibinfo{person}{George Karypis}.} \bibinfo{year}{2023}\natexlab{b}.
\newblock \showarticletitle{Mixed-Type Tabular Data Synthesis with Score-based Diffusion in Latent Space}.
\newblock \bibinfo{journal}{\emph{arXiv preprint arXiv:2310.09656}} (\bibinfo{year}{2023}).
\newblock


\bibitem[Zhang et~al\mbox{.}(2023a)]%
        {zhang2023generative}
\bibfield{author}{\bibinfo{person}{Tianping Zhang}, \bibinfo{person}{Shaowen Wang}, \bibinfo{person}{Shuicheng Yan}, \bibinfo{person}{Li Jian}, {and} \bibinfo{person}{Qian Liu}.} \bibinfo{year}{2023}\natexlab{a}.
\newblock \showarticletitle{Generative Table Pre-training Empowers Models for Tabular Prediction}. In \bibinfo{booktitle}{\emph{Proceedings of the 2023 Conference on Empirical Methods in Natural Language Processing}}. \bibinfo{pages}{14836--14854}.
\newblock


\bibitem[Zhang et~al\mbox{.}(2020)]%
        {zhang2020statistical}
\bibfield{author}{\bibinfo{person}{Yunjia Zhang}, \bibinfo{person}{Zhihan Guo}, {and} \bibinfo{person}{Theodoros Rekatsinas}.} \bibinfo{year}{2020}\natexlab{}.
\newblock \showarticletitle{A statistical perspective on discovering functional dependencies in noisy data}. In \bibinfo{booktitle}{\emph{Proceedings of the 2020 ACM SIGMOD International Conference on Management of Data}}. \bibinfo{pages}{861--876}.
\newblock


\bibitem[Zhao et~al\mbox{.}(2023)]%
        {zhao2023tabula}
\bibfield{author}{\bibinfo{person}{Zilong Zhao}, \bibinfo{person}{Robert Birke}, {and} \bibinfo{person}{Lydia Chen}.} \bibinfo{year}{2023}\natexlab{}.
\newblock \showarticletitle{Tabula: Harnessing language models for tabular data synthesis}.
\newblock \bibinfo{journal}{\emph{arXiv preprint arXiv:2310.12746}} (\bibinfo{year}{2023}).
\newblock


\bibitem[Zhao et~al\mbox{.}(2021)]%
        {zhao2021ctab}
\bibfield{author}{\bibinfo{person}{Zilong Zhao}, \bibinfo{person}{Aditya Kunar}, \bibinfo{person}{Robert Birke}, {and} \bibinfo{person}{Lydia~Y Chen}.} \bibinfo{year}{2021}\natexlab{}.
\newblock \showarticletitle{Ctab-gan: Effective table data synthesizing}. In \bibinfo{booktitle}{\emph{Asian Conference on Machine Learning}}. PMLR, \bibinfo{pages}{97--112}.
\newblock


\bibitem[Zheng et~al\mbox{.}(2023)]%
        {zheng2023dense}
\bibfield{author}{\bibinfo{person}{Lei Zheng}, \bibinfo{person}{Ning Li}, \bibinfo{person}{Xianyu Chen}, \bibinfo{person}{Quan Gan}, {and} \bibinfo{person}{Weinan Zhang}.} \bibinfo{year}{2023}\natexlab{}.
\newblock \showarticletitle{Dense Representation Learning and Retrieval for Tabular Data Prediction}. In \bibinfo{booktitle}{\emph{Proceedings of the 29th ACM SIGKDD Conference on Knowledge Discovery and Data Mining}}. \bibinfo{pages}{3559--3569}.
\newblock


\bibitem[Zhu et~al\mbox{.}(2022)]%
        {zhu2022permutation}
\bibfield{author}{\bibinfo{person}{Yujin Zhu}, \bibinfo{person}{Zilong Zhao}, \bibinfo{person}{Robert Birke}, {and} \bibinfo{person}{Lydia~Y Chen}.} \bibinfo{year}{2022}\natexlab{}.
\newblock \showarticletitle{Permutation-Invariant Tabular Data Synthesis}. In \bibinfo{booktitle}{\emph{2022 IEEE International Conference on Big Data (Big Data)}}. IEEE, \bibinfo{pages}{5855--5864}.
\newblock


\end{thebibliography}

\end{document}